\documentclass[accepted]{melba}

\usepackage{mwe} 

\usepackage{amsmath,amsfonts}

\usepackage{cite}
\usepackage{amsmath,amssymb,amsfonts}
\usepackage{graphicx}
\usepackage{textcomp}
\usepackage{url}
\usepackage{algorithm,algorithmicx}
\usepackage[noend]{algpseudocode}
\usepackage{textcomp}
\usepackage{multirow}
\usepackage{bm}
\usepackage{hyperref}
\usepackage{color}
\usepackage{rotating}
\usepackage{booktabs}
\usepackage{xr}
\usepackage{gensymb}
\usepackage{float}
\usepackage{subcaption}
\usepackage[normalem]{ulem}

\newcommand{\modelname}{KeyMorph}
\newcommand{\modelnames}{BrainMorph}

\newcommand{\moveimg}{\bm{x}_m}
\newcommand{\fiximg}{\bm{x}_f}
\newcommand{\alignimg}{\bm{x}_r}

\newcommand{\opttran}{\theta^*}
\newcommand{\norm}[1]{\left\lVert#1\right\rVert}
\newcommand{\nnparam}{\psi}
\newcommand{\nn}{f_\nnparam}
\DeclareMathOperator*{\argmax}{arg\,max}
\DeclareMathOperator*{\argmin}{arg\,min}
\newcommand{\bpara}[1]{\vspace{0.3cm} \noindent \textbf{#1}}
\usepackage{pifont}
\newcommand{\cmark}{\ding{51}}%
\newcommand{\xmark}{\ding{55}}%
\usepackage{colortbl} 
\newcommand{\rowcolorone}{\rowcolor[gray]{.9}}

\melbaid{2025:010}  
\doi{10.59275/j.melba.2025-59g7}
\melbaauthors{Wang et al.}  
\email{aw847@cornell.edu}
\volume{3}
\firstpageno{181}  
\melbayear{2025}  
\datesubmitted{2024-08-24}  
\datepublished{2025-05-27}  

\melbaspecialissue{Medical Imaging with Deep Learning (MIDL) 2020}
\melbaspecialissueeditors{Marleen de Bruijne, Tal Arbel, Ismail Ben Ayed, Hervé Lombaert}

\ShortHeadings{BrainMorph}{Wang et al.}

\title{BrainMorph: A Foundational Keypoint Model for Robust and Flexible Brain MRI Registration}

\author{
	\name Alan Q. Wang\aff{1,2},
	\name Rachit Saluja\aff{1,2},
    \name Heejong Kim\aff{1,2},
    \name Xinzi He\aff{1,2},
    \name Adrian Dalca\aff{3},
    \name Mert R. Sabuncu\aff{1,2}
}
\affiliations{
	\num 1 \addr Department of Radiology, Weill Cornell Medicine, New York City, NY, USA \\
    \num 2 \addr School of Electrical and Computer Engineering, Cornell University, Ithaca, NY, USA \\
	\num 3 \addr Computer Science and Artificial Intelligence Lab at the Massachusetts Institute of Technology, Cambridge, MA, USA
}

\abstract{
We present a keypoint-based foundation model for general purpose brain MRI registration, based on the recently-proposed \modelname~framework. 
Our model, called {\modelnames}, serves as a tool that supports multi-modal, pairwise, and scalable groupwise registration.
{\modelnames} is trained on a massive dataset of over 100,000 3D volumes, skull-stripped and non-skull-stripped, from nearly 16,000 unique healthy and diseased subjects.
\modelnames~is robust to large misalignments, interpretable via interrogating automatically-extracted keypoints, and enables rapid and controllable generation of many plausible transformations with different alignment types and different degrees of nonlinearity at test-time.
We demonstrate the superiority of {\modelnames} in solving 3D rigid, affine, and nonlinear registration on a variety of multi-modal brain MRI scans of healthy and diseased subjects, in both the pairwise and groupwise setting. 
In particular, we show registration accuracy and speeds that surpass many classical and learning-based methods, especially in the context of large initial misalignments and large group settings. 
All code and models are available at \texttt{https://github.com/alanqrwang/brainmorph}.
}

\keywords{Image registration, Multi-modal, Keypoint detection, Foundation model, Brain MRI}

\begin{document}

\twocolumn[\maketitle]

\section{Introduction}
\enluminure{R}{egistration} is a fundamental problem in biomedical imaging tasks. 
Multiple images, often reflecting a variety of contrasts, modalities, subjects, and underlying pathologies, are commonly acquired in many applications~\citep{uludaug2014general}. 
Registration seeks to spatially align these images in order to facilitate downstream analyses, like tracking longitudinal changes, studying disease progression, or analyzing population-level variability.

\begin{figure*}[t!]
\centering
\includegraphics[width=\textwidth]{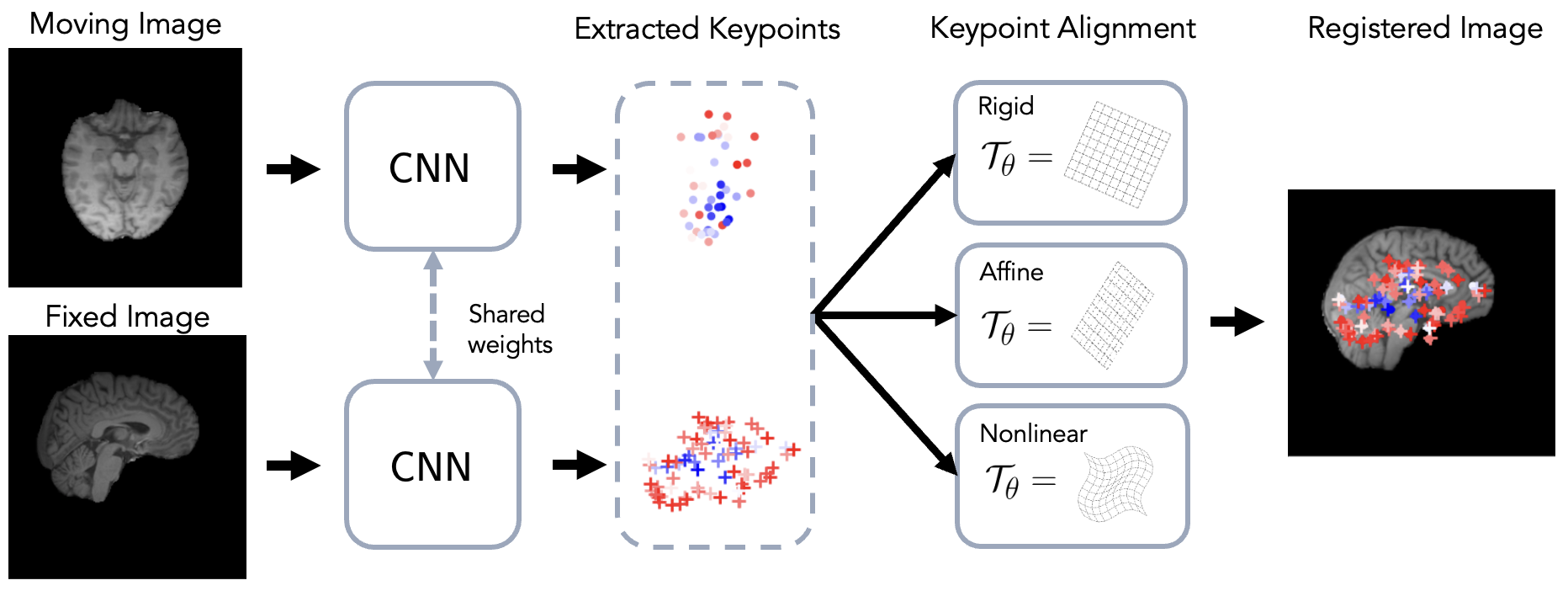}
   \caption{The framework of {\modelnames}. Fixed and moving 3D brain images are passed through the same keypoint detection network, which predicts $N$ keypoints useful for registration. The transformation parameters are then computed as a function of the keypoints, which are in turn used to resample the moving image. Keypoint colors denote depth (see Fig.~\ref{fig:imshow_points_ss_and_nss}).}
\label{fig:proposed_model}
\end{figure*}

Registration can be broken down into different types.
It may be performed within the same modality (unimodal) or across different modalities (multimodal).
Pairwise registration performs registration on an image pair, while groupwise registration performs registration on multiple images at once~\citep{Guyader2018Groupwise}. 
To perform the registration, various families of spatial transformations can be used, including rigid, affine, and nonlinear transformations.

Different lines of research have been explored to solve the registration task.
``Classical'' (i.e. non-learning-based) registration methods solve an iterative optimization of a similarity metric over a space of transformations, with additional regularization terms that restrict the space of plausible transformations~\citep{oliveira2014medical,sotiras2013survey}. 
Much research is dedicated to developing good transformations, similarity metrics, and optimization strategies.
While performant, these approaches are known to suffer from long run times, often requiring upwards of several minutes to register a pair of images.
In addition, these approaches are known to perform poorly when the initial misalignment between images is large (e.g. 90 degrees of rotational misalignment).

\begin{figure*}[t]
\centering
\includegraphics[width=\textwidth]{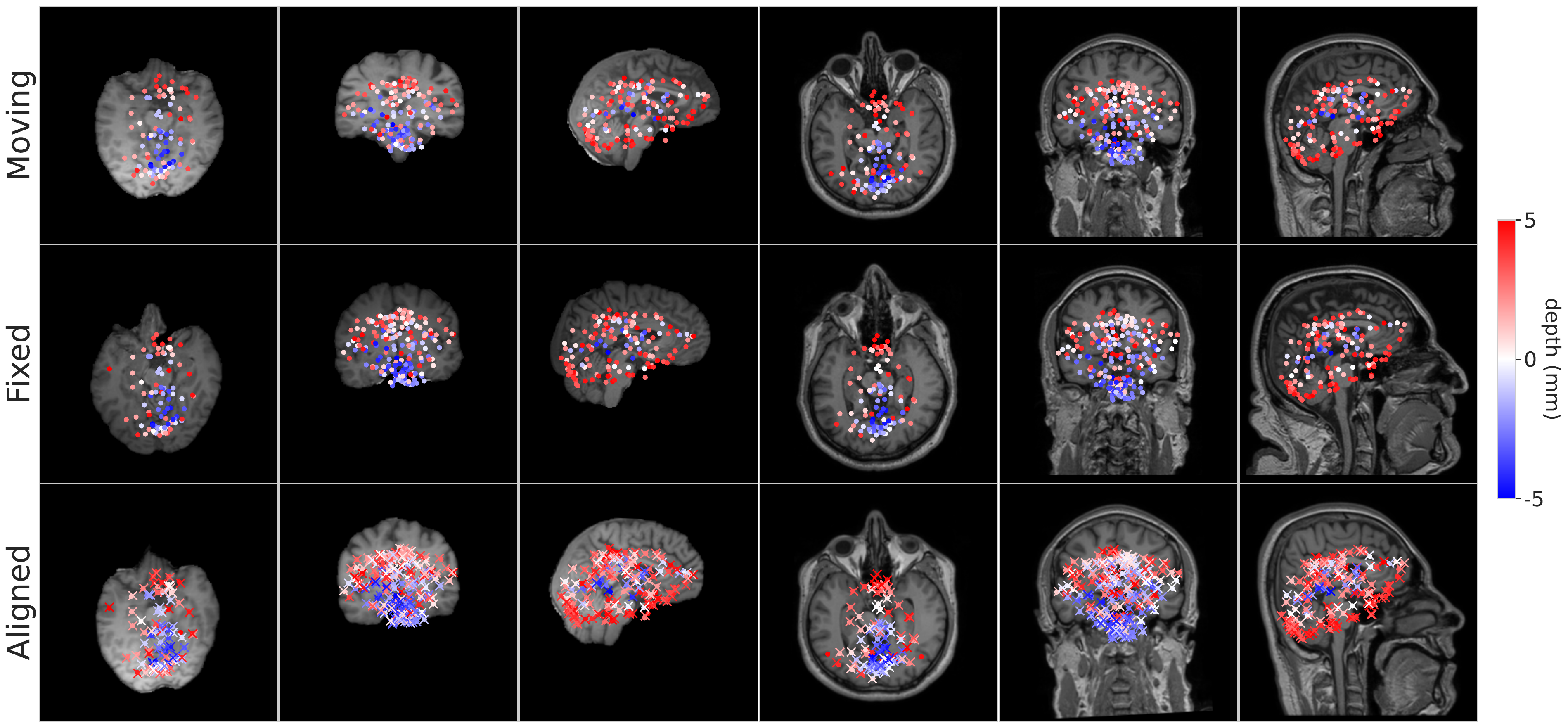}
   \caption{Moving, fixed, and aligned images in axial, sagittal, and coronal mid-slices.
   Keypoints extracted by {\modelnames} are overlaid. 
   For Aligned, aligned keypoints are dots and fixed keypoints are crosses.
   Color of the keypoint corresponds to depth.
   Note that keypoint weights are not visualized.}
\label{fig:imshow_points_ss_and_nss}
\end{figure*}

Another line of work decomposes the registration problem into two steps.
First, salient features (e.g. keypoints or contours) are extracted from images, and correspondences are established between the features of the image pair. 
Second, the transformation is derived which aligns these features and correspondences.
In this work, we refer to methods that extract keypoints as salient features as ``keypoint-based'' registration.
Largely, keypoint-based registration is advantageous in that the registration is relatively robust to initial misalignments, given good correspondences.
In addition, these methods enjoy superior interpretability, because the user can interrogate the correspondences which are driving the registration.
However, finding keypoints and establishing correspondences is a difficult task and is a subject of much research~\citep{lowe2004distinctive,Tang2023HD2Reg,Tang2019Neural}.

More recently, deep learning-based strategies have emerged which leverage large datasets of images to train a neural network to perform the registration task. 
These strategies use convolutional neural network (CNN) or transformer-based architectures~\citep{Ma2022Symmetric,Chen2021ViTVNet,Khan2021Transformers} that either output transformation parameters (e.g. rigid, affine or spline)~\citep{lee2019imageandspatial,de2019deep} or a dense deformation field~\citep{balakrishnan2019voxelmorph,hoffmann2022synthmorph} which aligns an image pair. 
These strategies are effective and are able to perform fast inference via efficient feed-forward passes.
However, they often fail when the initial misalignment is large and are usually less robust than their classical counterparts.
 
Recently, the {\modelname} framework proposed to combine the benefits of keypoint-based registration with deep learning using neural networks to automatically detect corresponding keypoints~\citep{yu2022keymorph,wang2023keymorph}.
Corresponding keypoints can then be used to compute the optimal transformation in closed-form, where the keypoints themselves are learned by a neural network.
Thus, {\modelname} may be seen as possessing all the benefits of keypoint-based registration, including robustness to large misalignments and better interpretability, while retaining the fast inference times of deep learning-based methods.
In addition, different transformation can be used according to user specifications, thereby enabling human controllability of the registration process.

In this work, we extend the KeyMorph framework into a general-purpose tool for brain MRI registration. 
Although the literature on deep-learning-based registration methods has been widely explored and many tools exist with support for brain MRIs~\citep{avants2008syn,konstantinos_ntatsis-proc-scipy-2023,hoffmann2022synthmorph}, most works focus on cross-subject pairwise registration on healthy subjects with skull-stripped images. 
Groupwise registration and support for non-skull-stripped and diseased subjects is often ignored, or is prohibitively slow or memory-intensive~\citep{konstantinos_ntatsis-proc-scipy-2023}.
Often, the tools that do exist require time-consuming preprocessing steps like skullstripping and pre-affine registration~\citep{balakrishnan2019voxelmorph,hoffmann2022synthmorph}.
In short, there is a lack of software tools for brain MRIs which is capable of supporting registration across a wide swath of use cases, including healthy and diseased subjects, pairwise/longitudinal/groupwise registration, and minimal assumptions on preprocessing (like skullstripping).

We refer to our proposed tool as {\modelnames}.
\modelnames{} is a tool built on the KeyMorph framework that is trained on a massive dataset of over 100,000 volumes consisting of both skull-stripped and non-skull-stripped data of diseased and normal subjects in a variety of MRI modalities.
Our tool supports rigid, affine, and nonlinear registration.
In addition to pairwise registration, we introduce a novel and memory-efficient approach to groupwise registration and demonstrate the superiority and scalability of our approach in population-level and longitudinal settings.
All code and models are available at: \\ \texttt{https://github.com/alanqrwang/brainmorph}.

This work builds on previous works on KeyMorph~\citep{yu2022keymorph,wang2023keymorph} in the following ways:
\begin{enumerate}
    \item We scale up training of the KeyMorph framework on a massive dataset of over 100,000 volumes at 1mm isotropic resolution, enabling our resulting model to be robust to full-resolution skull-stripped and non-skull-stripped data, a variety of MRI modalities, and diseased and normal subjects.
    \item We introduce a groupwise registration algorithm operating on the learned keypoints that is scalable and memory efficient.
    \item We provide code and all model variants such that our models can be used as a tool for the research community.
\end{enumerate}


\section{Background}
\bpara{Classical Methods.}
Pairwise iterative, optimization-based approaches have been extensively studied in medical image registration~\citep{hill2001medical,oliveira2014medical}. These methods employ a variety of similarity functions, types of deformation, transformation constraints or regularization strategies, and optimization techniques. 
Intensity-based similarity criteria are most often used, such as mean-squared error (MSE) or normalized cross correlation for registering images of the same modality~\citep{avants2009advanced, avants2008symmetric, hermosillo2002variational}. For registering image pairs from different modalities, statistical measures like mutual information or contrast-invariant features like MIND are popular~\citep{heinrich2012mind,hermosillo2002variational, hoffmann2020learning, mattes2003pet, viola1997alignment}.


\bpara{Keypoint-based Methods.}
Another registration paradigm first detects features or keypoints in the images, and then establishes their correspondence. This approach often involves handcrafted features~\citep{tuytelaars2008local}, features extracted from curvature of contours~\citep{rosenfeld1971edge}, image intensity~\citep{forstner1987fast, harris1988combined}, color information~\citep{montesinos1998differential, van2005boosting}, or segmented regions~\citep{matas2004robust, wachinger2018keypoint}. Features can be also obtained so that they are invariant to viewpoints~\citep{bay2006surf, brown2005multi,lowe2004distinctive, toews2013feature}. These algorithms then optimize similarity functions based on these features over the space of transformations~\citep{chui2003new,hill2001medical}. This strategy is sensitive to the quality of the keypoints and often suffer in the presence of substantial contrast and/or color variation~\citep{verdie2015tilde}. 

\bpara{Deep Learning-based Methods.}
In learning-based image registration, supervision can be provided through ground-truth transformations, either synthesized or computed by classical methods~\citep{cao2018deformable, dosovitskiy2015flownet, eppenhof2018pulmonary, lee2019image, uzunova2017training, yang2017quicksilver}. Unsupervised strategies use loss functions similar to those employed in classical methods~\citep{balakrishnan2019voxelmorph, dalca2019unsupervised, de2019deep, fan2018adversarial, krebs2019learning, qin2019, wu2015scalable, hoopes2021hypermorph}. Weakly supervised models employ (additional) landmarks or labels to guide training~\citep{balakrishnan2019voxelmorph, fan2019birnet, hu2018label, hu2018weakly}.


Recent learning-based methods compute image features or keypoints~\citep{ma2021image,moyer2021equivariant} that can be used for image recognition, retrieval, tracking, or registration. 
Learning useful features or keypoints can be done with supervision~\citep{verdie2015tilde,yi2016lift,yi2018learning}, self-supervision~\citep{detone2018superpoint,liu2021same} or without supervision~\citep{barroso2019key, lenc2016learning, ono2018lf}. 
Finding correspondences between pairs of images usually involves identifying the learned features which are most similar between the pair.
In contrast, our method uses a single network which extract/generates corresponding keypoints directly from both the moving and fixed image. 
The keypoints between the moving and fixed image are guaranteed to be in correspondence, since one network extracts keypoints from both images.
We optimize these corresponding keypoints directly for the registration task (and not using any intermediate keypoint supervision).

\begin{table*}[t]
\scriptsize
\centering
\begin{tabular}{lcccc}
\toprule
\textit{Model}      & \textit{CPU Time, pre-processed} & \textit{GPU Time, pre-processed} & \textit{CPU Time, raw} & \textit{GPU Time, raw} \\ \midrule 
\rowcolorone
ANTs, Rigid         & 101.38$\pm$2.33 & -               & 143.38$\pm$2.33 & -          \\
ANTs, Affine        & 110.45$\pm$2.94  & -               & 142.45$\pm$2.94 & -          \\
\rowcolorone
ANTs, Syn           & 216.03$\pm$3.14 & -       & 248.03$\pm$3.14 & -                \\
\modelnames{}, Rigid   & 109.84$\pm$1.80               & \textbf{1.05$\pm$0.29}  & 108.27$\pm$1.82             &   \textbf{1.05$\pm$0.25}           \\ 
\rowcolorone
\modelnames{}, Affine   & 109.63$\pm$1.84               & \textbf{1.04$\pm$0.36}  & 110.31$\pm$1.90             &    \textbf{1.05$\pm$0.32}         \\ 
\modelnames{}, TPS  & 180.14$\pm$1.99               & \textbf{1.24$\pm$0.30}  & 180.40$\pm$1.91             &    \textbf{1.25$\pm$0.31}          \\ \bottomrule
\end{tabular}%
\caption{
Average computation time in seconds for pairwise registration across different models. 
We separate between times when performing registration with pre-processed data and raw data.
For ANTs, pre-processing requires skull-stripping with HD-BET, which adds an additional 32 seconds~\citep{isensee2019hdbet}.
For SynthMorph, pre-processing requires an initial robust affine registration into a reference space.
BrainMorph timings are based on BrainMorph-L.
Note: all times are inference times. BrainMorph requires about 5 days of training time (compute details in Sec.~\ref{sec:training_details}), while ANTs does not.}
\label{tab:timing}
\end{table*}



Learning-based methods for multi-modal registration are of great practical utility and often-studied in the literature.
Most works require, in addition to the moving and fixed image, a corresponding image in a standard space which can be compared and which drives the alignment, usually in the form of segmentations.
\citep{zhang2022twostep} address multi-modal retinal images and handle multi-modality by transforming each image to a standard grayscale image via vessel segmentation. 
A standard feature detection and description procedure is used to find correspondences from these standard images.
Other works \citep{song2022crossmodal} rely on segmentations from ultrasound and magnetic resonance images to align them.
Obtaining these segmentations may be costly, add additional computational complexity to the registration procedure, or be specific to the anatomies/modalities in question.
In contrast, our method can be applied generally to any registration problem.
In addition, we present a variant of our model which only relies on the images themselves during training.
In our experiments, we find that this variant outperforms state-of-the-art baselines while also performing comparably to a variant of our model which leverages segmentations.

\section{{\modelnames}}
Our tool, {\modelnames}, is based on the previously published {\modelname} framework and we refer the reader to prior papers for more details~\citep{yu2022keymorph,wang2023keymorph}.
Let~$(\moveimg, \fiximg)$ be a moving (source) and fixed (target) image\footnote{Although we consider 3D volumes in this work,~\modelnames~is agnostic to the number of dimensions. The terms ``image'' and ``volume''  are used interchangeably.} pair, possibly of different contrasts or modalities.  
Additionally, we denote by $\mathcal{T}_{\theta}$ a parametric coordinate transformation with parameters $\theta$, such as those discussed in~\ref{sec:transformations}. 
The goal is to find the optimal transformation $\mathcal{T}_{\theta^*}$ such that the registered image~$\alignimg = \moveimg \circ \mathcal{T}_{\theta^*}$ aligns with some fixed image~$\fiximg$, where~$\circ$ denotes the spatial transformation of an image.

{\modelnames} works by detecting $N$ keypoints $\bm{P}\in \mathbb{R}^{D\times N}$ from any given image.
In this work, $D=3$.
The keypoints are detected by a neural network $\nn$.
Since a single $\nn$ detects keypoints for any image and the detected keypoints are used to optimize a registration objective (see Section~\ref{sec:training}), the keypoints for any arbitrary image pair are in correspondence.
Given corresponding keypoint sets $\bm{P}$ and $\bm{Q}$, the optimal transformation can be derived using a keypoint solver, which outputs the optimal transformation parameters as a function of the keypoints: $\opttran(\bm{P}, \bm{Q}, \bm{w})$.
Here, we optionally include as arguments a vector of weights $\bm{w}$ which weight the correspondences, such that lower weights lead to a lower contribution to the overall alignment.
Further details are given in Section~\ref{sec:model_details}.

Fig.~\ref{fig:proposed_model} depicts a graphical overview of {\modelnames}.
Note this formulation unlocks the benefits of keypoint-based registration, including robustness to large misalignments and interpretability (as compared to other learning-based methods) via visualizing the keypoints. 
Moreover, this formulation enables controllability in the sense that different transformations can be used to align the keypoints. 
In particular, during training, this allows for heterogeneity in training, such that the model can be robust to a wide variety of transformation types.
At test-time, one can generate a dense set of registrations; the controllable nature of this framework enables the user to select the preferred registration.

\subsection{Keypoint Detection Network}
{\modelnames} can leverage any deep learning-based keypoint detector~\citep{ma2021image,detone2018superpoint,barroso2019key}.
In this work, we use a UNet-style architecture which takes in a single channel image and outputs $N$ feature maps.
We are interested in preserving translation equivariance; to this end, we leverage a center-of-mass (CoM) layer~\citep{ma2020volumetric, sofka2017fully} as the final layer, which computes the center-of-mass for each of the $N$ activation maps. 
This specialized layer is (approximately) translationally-equivariant and enables precise localization. 
Since the CoM layer expects positive values at every grid location, we insert a ReLU activation before the CoM layer.

\subsection{Training}
\label{sec:training}
Training {\modelnames} involves optimizing the learnable parameters within the CNN $\nn$ for pairwise registration.
During training, we randomly sample pairs of moving and fixed images, and the general objective is:
\begin{align}
\begin{split}
\argmax_\nnparam \ &\mathbb{E}_{(\moveimg, \fiximg)} \  \mathcal{L}_{sim}\left(\moveimg \circ \mathcal{T}_{\theta^*}, \fiximg\right)  \\
&\text{where } \ \ \theta^* = \opttran\left(\nn(\fiximg), \nn(\moveimg), \bm{w} \right)
\label{eq:objective}
\end{split}
\end{align}
where $\mathcal{L}_{sim}(\cdot, \cdot)$ measures image similarity between its two inputs.
Weights $\bm{w}$ for keypoints correspondences are found as follows.
First, we compute the energy (i.e. the aggregated sum) of each of the $N$ activation maps for both the fixed and moving image.
Then, we multiply the corresponding energies, and compute the softmax of the $N$ energies to arrive at normalized weights~\citep{moyer2021equivariant}.

In this work, we choose transformations whose optimal parameters can be solved in a closed-form and differentiable manner so that $\nn$ can be trained in an end-to-end fashion. 
Thus, the neural network is incentivized to detect $N$ anatomically-consistent keypoints from a given image, such that a good registration can be achieved. 
Note that we do not rely on any ground truth keypoints as supervision.

The {\modelnames} framework enables flexibility in training depending on the choice of the loss function and the transformation used.
$\mathcal{L}_{sim}$ can be any similarity function and can vary during training depending on the image pairs.
In this work, we use MSE or Dice loss, depending on the current image pair.
The closed-form optimal solution $\theta^*$ can depend on a hyperparameter $\lambda$, such as in TPS, which can be set to a constant or sampled from a distribution $\lambda \sim p(\lambda)$ during training.
More details on the training details we used in this work are presented in Section~\ref{sec:training_details}.
Once the model is trained, it can be used for both pairwise and groupwise registration during inference time, as described below. 

\subsection{Pairwise Registration}
Pairwise follows straightforwardly from the training setup.
Given a fixed image $\fiximg$ and moving image $\moveimg$, {\modelnames} performs pairwise registration as follows:
\begin{align}
\begin{split}
\bm{x}_r = \moveimg \circ \mathcal{T}_{\theta^*} \ \text{where} \ \theta^* = \opttran\left(\nn(\fiximg), \nn(\moveimg) \right)
\end{split}
\end{align}

Note that at test time, we can use any transformation $\mathcal{T}_\theta$ (e.g, rigid, affine or TPS with any hyperparameter value), which would yield a different alignment based on the same keypoints. 

\subsection{Groupwise Registration}
Groupwise registration methods try to mitigate uncertainties associated with any one image by simultaneously registering all images in a population. This incorporates all image information in the registration process and eliminates bias towards a chosen reference frame~\citep{toews2013featurebased,AGIER2020101564}.

We propose a novel algorithm for groupwise registration based on detected keypoints.    
In the groupwise setting, we have $M$ subjects to align; thus we are solving for the optimal transformation as well as the optimal average space simultaneously.
To achieve this with $N$ keypoints $\bm{P} \in \mathbb{R}^{D \times N}$ per subject, we optimize for the average space and the optimal transformations in an iterative, coordinate-ascent strategy by alternating the following two steps:
\begin{enumerate}
    \item Given points $\bm{P}_k$ at the current iteration $k$, compute the average keypoints $\bar{\bm{P}}_{k+1}$
    \item Compute new points $\bm{P}_{k+1}$ by registering all points $\bm{P}_k$ to $\bar{\bm{P}}_{k+1}$
\end{enumerate}

After $K$ iterations, the algorithm converges to the average space $\bar{\bm{P}}_K$, and the final registration for the $i$'th image is found by transforming the image according to the transformation that aligns points $\bm{P}_0$ to $\bm{P}_K$.

Note that this algorithm only relies on keypoints.
Thus, it is computationally efficient since keypoints can be precomputed and done serially, whereas other works must fit all images in memory at once.
In our experiments, we demonstrate the scalability of our approach by registering more than 100 volumes simultaneously.

\section{Materials and Methods}
\subsection{Dataset}\label{sec:dataset}
We train {\modelnames} on a massive dataset of over 100,000 images from nearly 16,000 unique subjects.
All datasets are gathered from publicly-available brain studies, datasets, and challenges.
The full list of datasets is given in the Appendix.
Our tool requires the following two widely-used pre-processing steps for all image inputs: resampling to 1mm isotropic, cropping/padding to 256x256x256.
Min-max rescaling to [0, 1] is performed as the first layer in the network, and thus we do not consider it a pre-processing step.

For training, we reorient all brains to MNI space\footnote{During training, we apply random affine transformations as an augmentation strategy.}.
For purposes of obtaining training data of both non-skullstripped and skullstripped data, 
we perform skull-stripping on all images with HD-BET~\citep{isensee2019hdbet}, a robust deep learning-based skull-stripping tool. 
For images without extreme lesions, we further generate segmentations with SynthSeg~\citep{billot2020learning,billot_synthseg_2023}, which produces parcellations of 33 brain regions.
We do not perform segmentation on images with extreme lesions.
See the Appendix for the full list of brain regions.
Datasets used for training and testing are non-overlapping, to ensure no data leakage occurs. Further details on evaluation datasets are provided in Sec.~\ref{sec:evaluation_datasets} and the Appendix.

\subsection{Training Details}\label{sec:training_details}
We are interested in learning foundational keypoints for the end goal of general-purpose brain registration.
Thus, the keypoints should be optimized such that they are robust to a variety of brain MRI modalities and transformation types (rigid, affine, and nonlinear).
Note that, {\modelnames} is amenable to a variety of training strategies (pairwise sampling, loss function, and transformation type).
We would like the network to be able to handle uni-modal, multi-modal, and longitudinal image pairs, with transform types including rigid, affine, and TPS.
To do so, we perform heterogeneous training with different tasks for a single foundation model, where the task is randomly sampled in each mini-batch.

During training, we use two loss types: Dice of segmentation labels and mean-squared-error (MSE) of pixel values.
We use three transformation types: rigid, affine, and TPS (nonlinear).
TPS has a hyperparameter $\lambda$ which controls the degree of nonlinearity.

\begin{table}[tb]
\footnotesize
    \caption{Summary of \modelnames{} training.
    }\label{tab:keymorph_training}
    \centering
    \begin{tabular}{l@{\quad}c@{\quad}cccc}
    \toprule
    \textit{Image pair} & \textit{Transform type} & \textit{Loss} \\
    \midrule
    \rowcolorone
    Normal & TPS & Dice \\
    Skullstripped, lesion & Affine  & MSE \\
    \rowcolorone
    Skullstripped, longitudinal               & Rigid & MSE  \\
    \bottomrule
    \end{tabular}
\end{table}

In general, we are constrained by the following rules.
\begin{enumerate}
    \item 
To use MSE loss, we must sample skull-stripped same-modality pairs.
\item
For Dice loss, we may sample pairs which have corresponding segmentations (this precludes brains with lesions, for which SynthSeg~\citep{billot2020learning,billot_synthseg_2023} cannot reliably segment).
\item
For longitudinal image pairs, we use rigid transformation to simulate realistic downstream usage.
\item
For image pairs with lesions, we use a restrictive affine transformation, as TPS will not guarantee bijective correspondence between images.
\end{enumerate}
Table~\ref{tab:keymorph_training} summarizes the training strategy used, which we choose according to the above constraints.
At every training iteration, we sample uniformly across the three image pair types.
Thus, the model is trained to optimize the registration performance across all three image tasks with equal weighting.

We experimented with $N$ = 128, 256, and 512 keypoints, and perform a thorough analysis of the relationship between number of keypoints and registration performance in Section~\ref{sec:num_keypoints}.
Weights $\bm{w}$ are applied for all models during training and evaluation.
For TPS transformations, we sample $\lambda$ during training from a log-uniform distribution $p(\lambda) = \text{LogUnif}(0, 10)$.
During testing, we can choose a $\lambda$ value that lies in the support of this distribution.
In addition, during training, in each mini-batch, we compute TPS on 32 keypoints chosen uniformly at random~\citep{donato2002approximatetps}.
This is because TPS has a high memory requirement due to computing pairwise distances between every keypoint and grid location.
Note that at test-time, we compute TPS on the full set of keypoints, and compute pairwise distances in a chunk-wise, iterative fashion to bypass this memory requirement.
When minimizing Dice loss, we sample 14 regions uniformly at random for computational purposes.

For all models, we used a batch size of 1 image pair and the Adam optimizer~\citep{kingma2017adam} for training.
We train for a total of 160K gradient steps.
The following uniformly-sampled augmentations were applied to the moving image across all dimensions during training: rotations $[-180\degree, +180\degree]$, translations $[-30, 30]$ voxels, scaling factor $[0.8, 1.2]$, and shear $[-0.1, 0.1]$.
All training and GPU testing was performed on a machine equipped with an AMD EPYC 7513 32-Core processor and an Nvidia A100 GPU. CPU testing was performed on a machine equipped with an Intel Xeon Gold 6330 CPU @ 2.00GHz. 
All {\modelnames} models are implemented in PyTorch. 

\subsection{Model Details}\label{sec:model_details}
\begin{table}[tb]
\footnotesize
    \caption{\modelnames{} backbone variants. Model size measured in megabytes (MB). \# downsampling denotes the number of downsampling layers.}
    \label{tab:backbone_variants}
    \centering
    \begin{tabular}{l@{\quad}c@{\quad}cccc}
    \toprule
     & \textit{\# parameters} & \textit{\# downsampling} & \textit{model size} \\
    \midrule
    \rowcolorone
    {\modelnames}-S      & 4M  & 4  & 48MB \\
    {\modelnames}-M   & 16M & 5  & 196MB\\
    \rowcolorone
    {\modelnames}-L  & 66M & 6 & 791MB \\
    \bottomrule
    \end{tabular}
\end{table}

Our architecture backbone consists of a truncated UNet, which is identical to a standard UNet except all layers which operate at the original resolution (e.g. after the last upsampling layer) are removed~\citep{ulyanov2016instance}. 
All truncated UNets we use have two convolutional blocks at each resolution.
Thus, the final center-of-mass layer extracts keypoints at half-resolution.
This enables us to train deeper networks with a bottleneck operating on a very coarse grid, which we empirically find leads to better performance.
In particular, we report results on three variants of the truncated UNet, which differ in the capacity as a function of the number of downsampling layers.
We refer to them as {\modelnames}-S, {\modelnames}-M, and {\modelnames}-L, and summarize them in Table~\ref{tab:backbone_variants}.
All references to {\modelnames} are {\modelnames}-L models, unless otherwise noted.

\subsection{Self-supervised Pretraining}
We employ the following self-supervised pre-training strategy to aid in keypoint detector initialization, essentially encouraging equivariance of the keypoint extractor with respect to affine image deformations. 
Note that past works have leveraged equivariant networks~\citep{billot2023se3equivariant}; however, we find these networks unstable to train and lack the capacity to capture variability present in our large datasets.
Using a single subject, we pick a random set of keypoints~$\bm{P}_0$ by sampling uniformly over the image coordinate grid. 
During pre-training, we apply random affine transformations to the input image as well as~$\bm{P}_0$, and minimize the following keypoint loss:
\begin{equation}
\argmin_\nnparam \sum_i \mathbb{E}_{\bm{A}}\norm{\bm{A}\bm{P}_0 - \nn\left(\bm{x}^{(i)} \circ \bm{A}\right)}^2_2. 
\end{equation} 
Here, $\bm{A}$ is an affine transformation drawn from a uniform distribution over the parameter space. 

We train for a total of 480K gradient steps.
We use the same augmentation strategy as training, except that we linearly increase the degree of augmentation such that maximum augmentation is reached after 160K gradient steps.
We use the same dataset for training and pretraining. 
Note that we assume that all the training data are in the same orientation and roughly in the center of the image in order for the sampled keypoints to apply well to all images in the dataset.

%

\begin{figure*}[t]
\centering
\includegraphics[width=\linewidth]{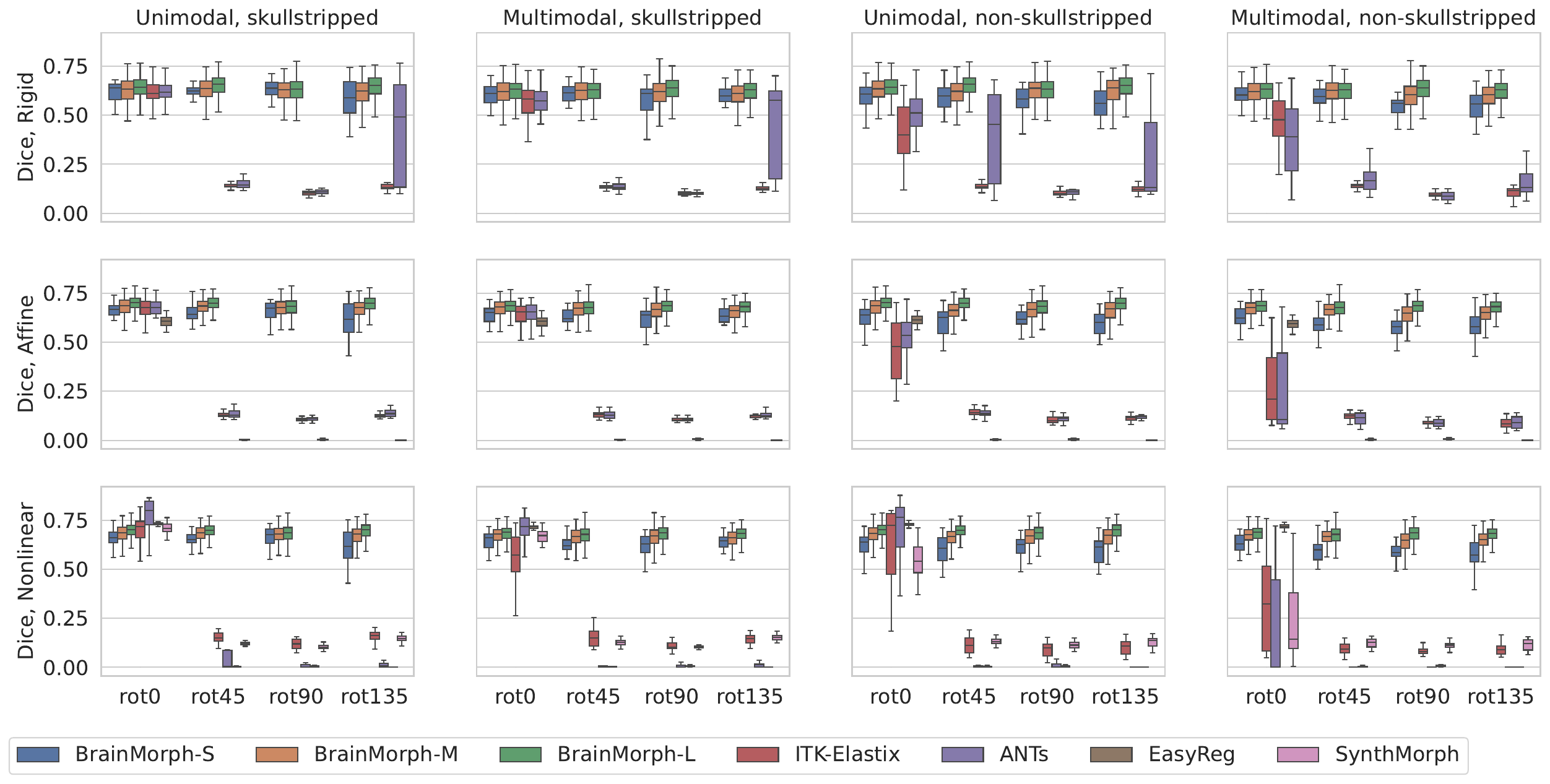}
\caption{Dice performance on pairwise registration. Higher is better. Unimodal/multimodal, skull-stripped/non-skull-stripped.}
\label{fig:lineplot_harddice_vs_rot_ss_nss_unimodal_multimodal}
\end{figure*}

\begin{figure*}[h!]
\centering
\includegraphics[width=\linewidth]{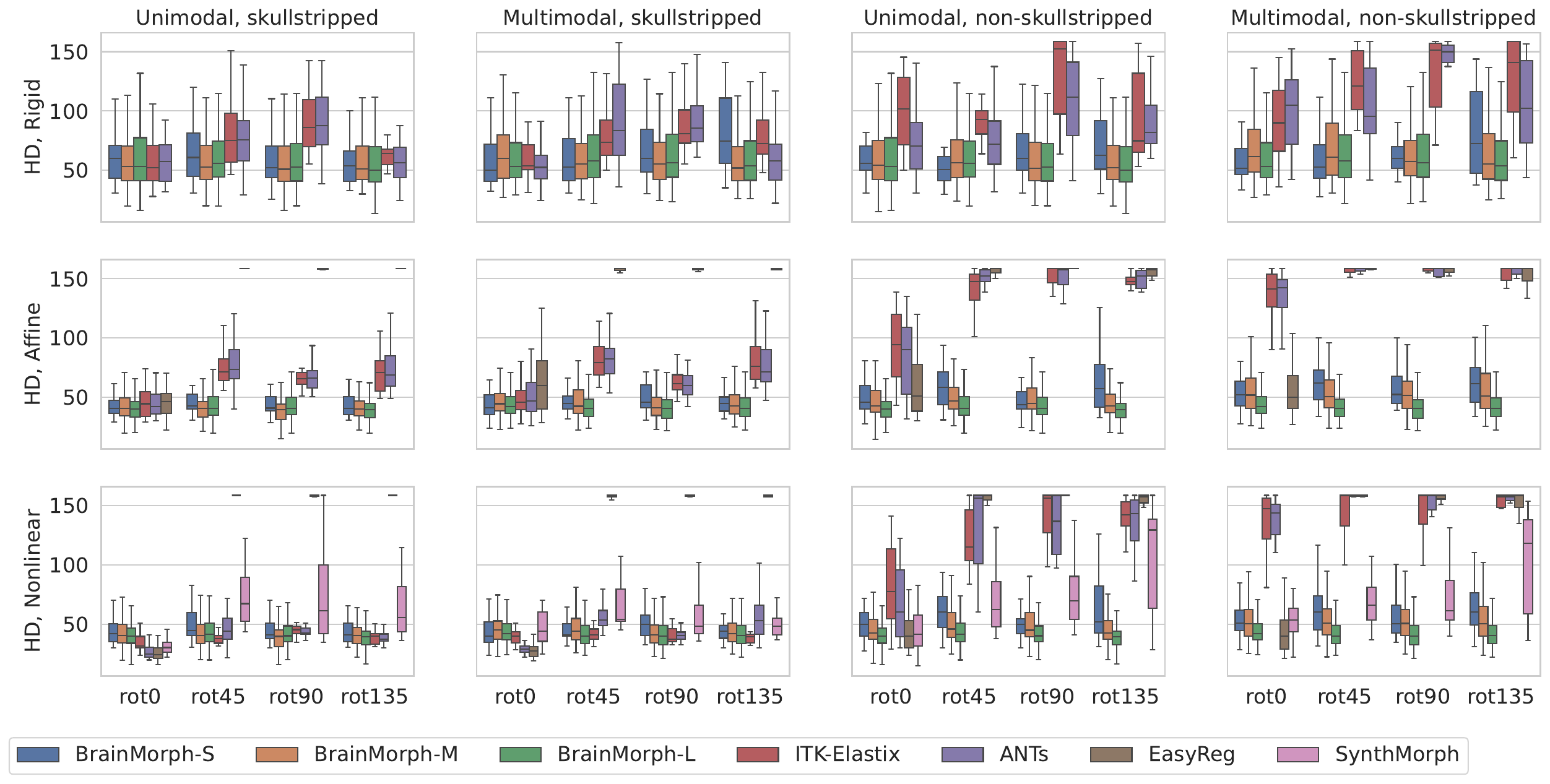}
\caption{HD performance on pairwise registration. Lower is better. Unimodal/multimodal, skull-stripped/non-skull-stripped.}
\label{fig:lineplot_hausd_vs_rot_ss_nss_unimodal_multimodal}
\end{figure*}

\section{Experimental Setup}\label{sec:experiments}

\begin{figure*}[ht]
\centering
\includegraphics[width=\linewidth]{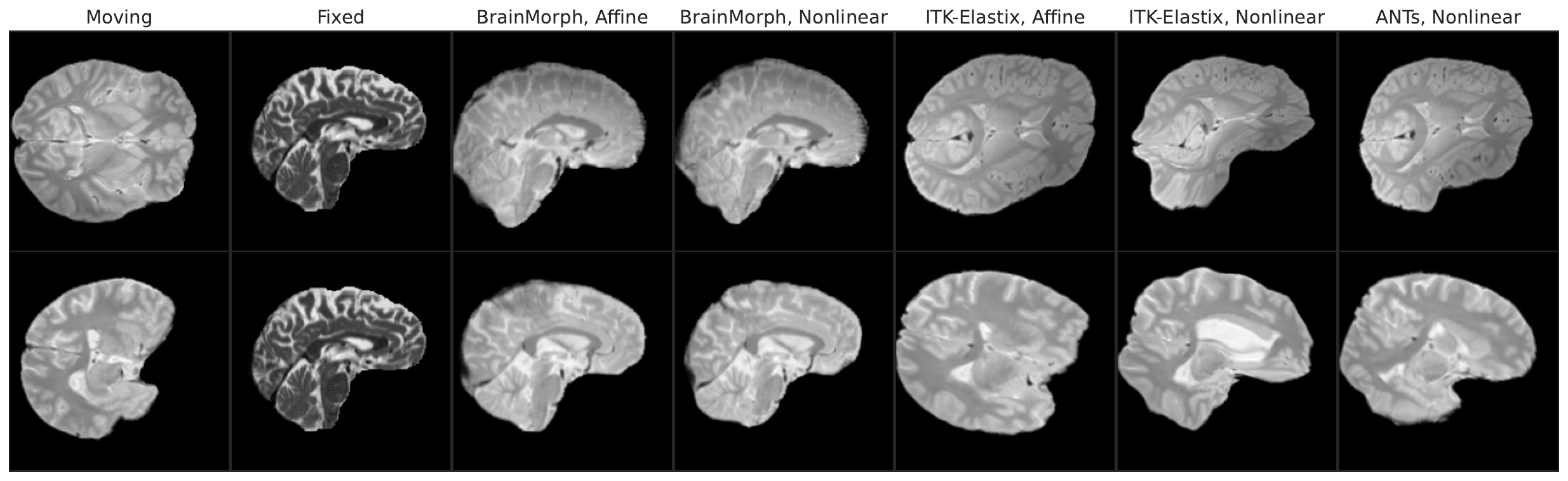}
   \caption{Pairwise registration results for \modelnames{} and selected baselines. In first row, a 90 degree rotation is applied. In second row, a 135 degree rotation is applied.}
\label{fig:imshow_pairwise}
\end{figure*}

\begin{figure*}[h!]
\centering
\includegraphics[width=0.8\linewidth]{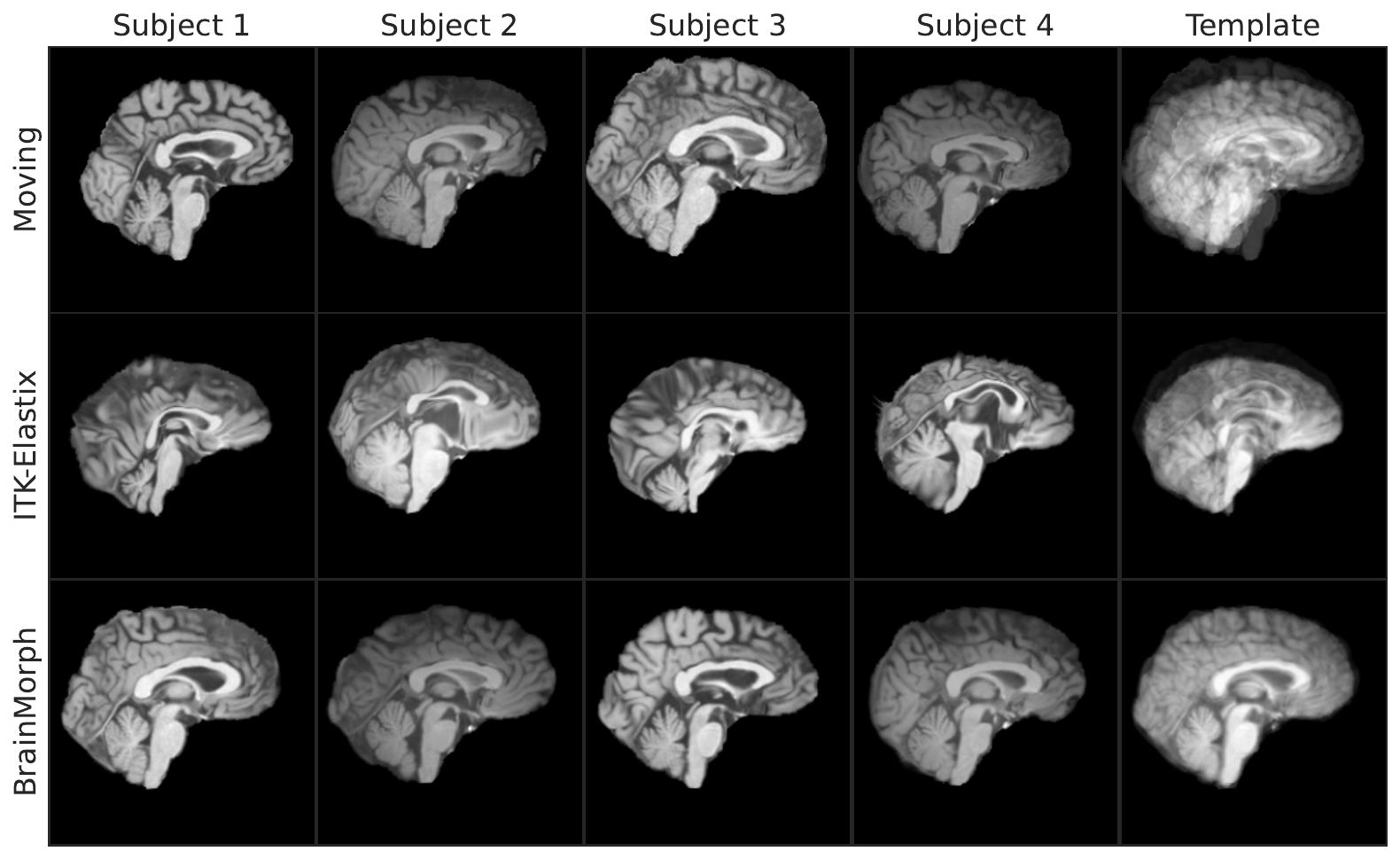}
   \caption{Groupwise registration with group size of 4. For ITK-Elastix, B-spline registration is used. For {\modelnames}, TPS with $\lambda=0$ is used. The last column shows the average brain in the optimized template space.}
\label{fig:imshow_groupwise}
\end{figure*}

\subsection{Evaluation Datasets}\label{sec:evaluation_datasets} 
We use the following datasets for evaluation of all models. Note that these datasets are not included in training.
\begin{enumerate}
\item 
For healthy pairwise and cross-subject groupwise experiments, we evaluate on the IXI brain MRI dataset\footnote{\url{https://brain-development.org/ixi-dataset/}}. 
Each subject has T1, T2, and PD-weighted 3D MRI scans in spatial alignment, so we can use this dataset for both unimodal and multimodal registration experiments. 
We evaluate on 100 subjects. 
\item 
For diseased subjects with lesions, we evaluate on the test split of the RSNA-ASNR-MICCAI BraTS dataset (to ensure no dataset contamination), which consists of adult brains with gliomas acquired with T1, T1gd, T2, and FLAIR sequences\footnote{\url{https://www.rsna.org/rsnai/ai-image-challenge/brain-tumor-ai-challenge-2021}}.
We also evaluate on a dataset of brains with multiple sclerosis~\citep{Muslim2022msdataset}, which consists of 60 MS patients with T1, T2, and FLAIR sequences.
\item 
For longitudinal experiments, we evaluate on the OASIS2 dataset, which consists of longitudinal MRI brains in nondemented and demented older adults\footnote{\url{https://www.oasis-brains.org/}}.
The number of timepoints per subject ranges between 2 and 10.
\end{enumerate}
For all evaluation datasets, we perform resampling to 1mm isotropic and cropping/padding to $256^3$ image size, which is standard across all baselines.

We use a pre-trained and validated SynthSeg model~\citep{billot2020learning} to automatically delineate 23 regions of interest (ROIs)\footnote{ROIs were pallidum, amygdala, caudate, cerebral cortex, hippocampus, thalamus, putamen, white matter, cerebellar cortex, ventricle, cerebral white matter, and brainstem.}. 
Furthermore, all performance evaluations were based on examining the overlap of ROIs in the test images.

\begin{figure*}[t]
\centering
\includegraphics[width=0.9\linewidth]{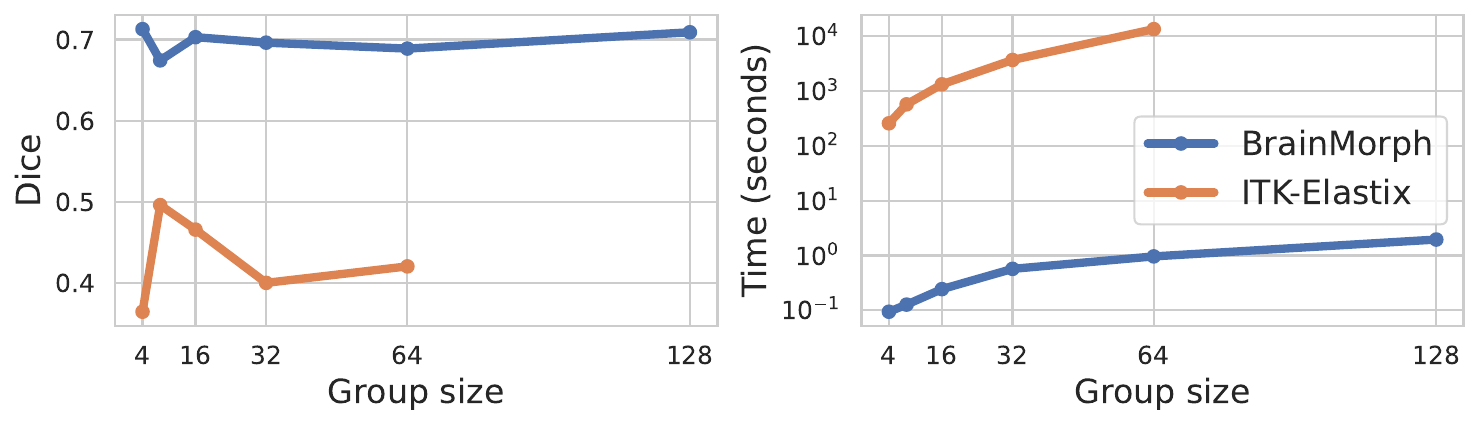}
   \caption{Groupwise registration using \modelnames{} and ITK-Elastix. For \modelnames{}, the transformation used is TPS with $\lambda=0$. For ITK-Elastix, the transformation is nonlinear b-spline. We observe significant improvement in groupwise registration capabilities of \modelnames{}, at a time reduction of nearly 4 orders of magnitude. This scalability and time reduction is possible due to precomputing keypoints for all images in the group, and performing the groupwise registration algorithm on keypoints directly.}
\label{fig:lineplot_dice_time_vs_groupsize}
\end{figure*}

\subsection{Test-time Performance Evaluation}
For pairwise experiments, we use each test subject as a moving volume~$\moveimg$, paired with another random test subject treated as a fixed volume~$\fiximg$. 
We simulate different degrees of misalignment by transforming~$\moveimg$ using rotation. 
Rotation is applied to all 3 axes at the specified degree. 
We use the predicted transformation to resample the moved segmentation labels on the fixed image grid. 
Unimodal/multimodal registration is an independent variable in our experiments.
We experiment with rigid, affine, and nonlinear registration types for all models and baselines.

For longitudinal experiments, we perform groupwise registration on all available timepoints, and restrict to rigid transformations only.
For cross-subject groupwise experiments, we sample different subject, same-modality images and experiment with varying group sizes in [4, 8, 16, 32, 64, 128]. 
We restrict to nonlinear transformations only.
Similar to pairwise experiments, we simulate different degrees of misalignment by transforming all images using rotation applied to all 3 axes at the specified degree. 

\subsection{Metrics}
For all experiments, we quantify alignment quality and properties of the transformation using Dice overlap score and Hausdorff distance (HD). 

\begin{figure*}[t!]
    \centering
    \begin{subfigure}[t]{0.45\textwidth}
        \centering
        \includegraphics[height=1.75in]{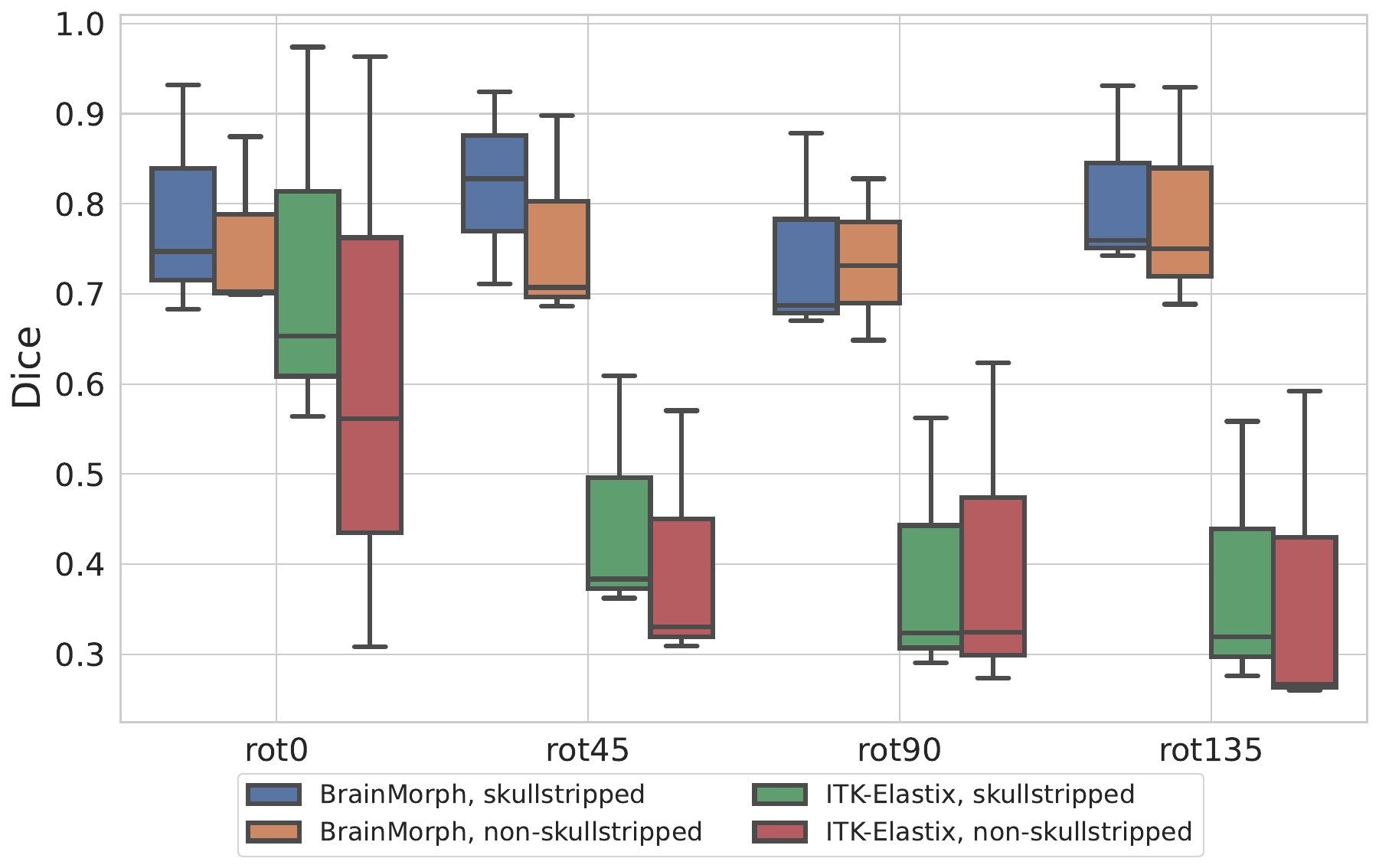}
        \caption{}
        \label{fig:boxplot_longitudinal}
    \end{subfigure}%
    ~ 
    \begin{subfigure}[t]{0.45\textwidth}
        \centering
        \includegraphics[height=1.75in]{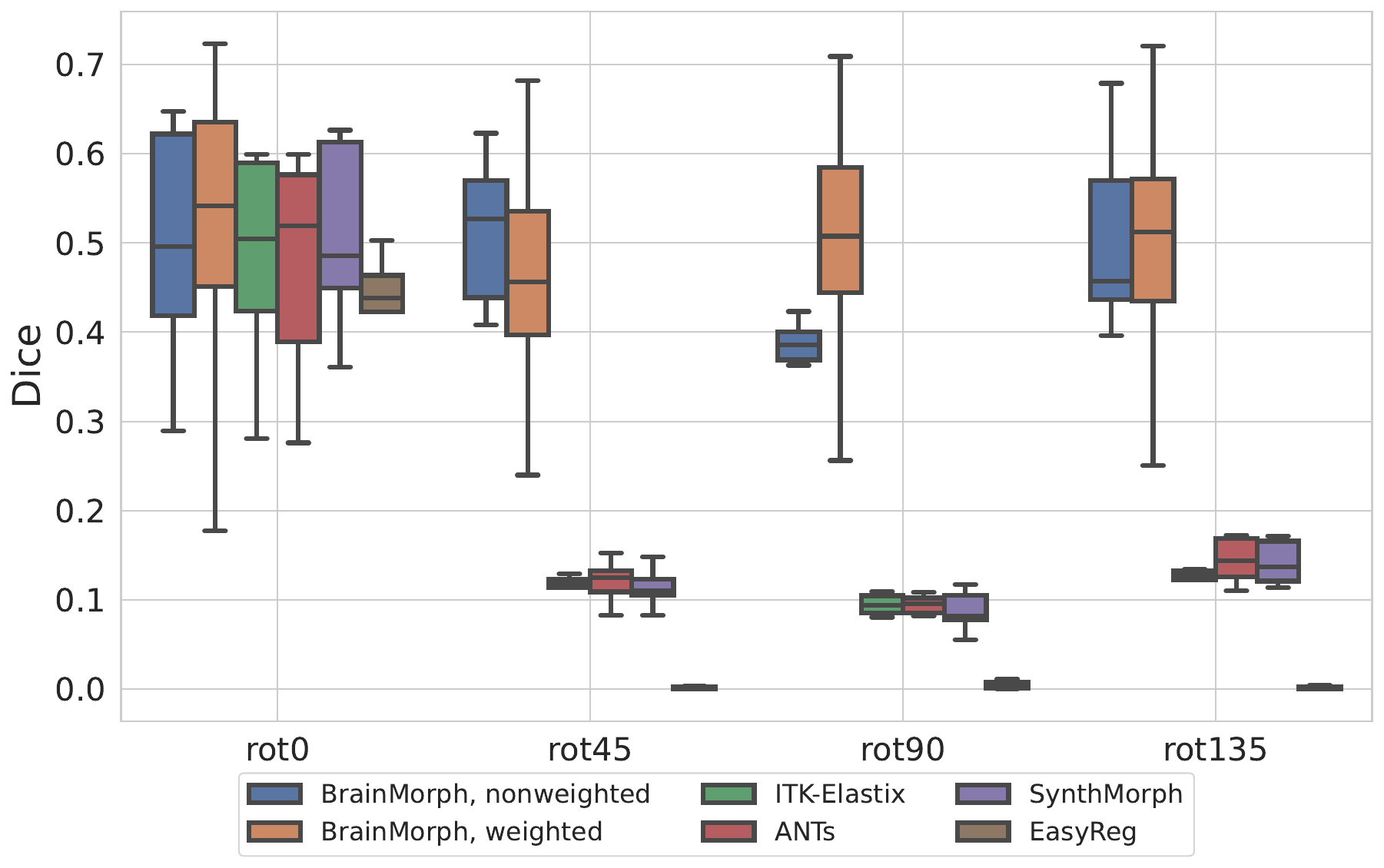}
        \caption{}
        \label{fig:boxplot_lesion_affine}
    \end{subfigure}
    \caption{a) Registration performance on longitudinal registration. Rigid transformations for all models. b) Registration performance on subjects with lesions. Affine transformations for all models, except for SynthMorph which only supports dense.}
\end{figure*}

\subsection{Baselines}\label{sec:baselines}
As the goal of this work is to develop a general-purpose tool for brain MRI registration, we wish to compare our proposed model against state-of-the-art and easily-accessible tools for this purpose. 
Our intended users are practitioners who desire a simple, easy-to-use tool that performs registration with minimal pre-processing or setup.
Thus, we adhere to all instructions required by baselines (including intensity normalization, initial robust registration, etc.), and do not assume that data are pre-processed. 
Note that since skullstripping vs. non-skullstripping is an independent variable in our experiments, we do not perform any skullstripping as part of any baseline's preprocessing requirements.

\begin{itemize}
    \item 
\textbf{ITK-Elastix} is a widely-used software package which supports  pairwise and groupwise registration~\citep{konstantinos_ntatsis-proc-scipy-2023}.
Rigid, affine, and bspline alignments are supported.
For all registrations, we perform a multi-resolution pyramid strategy at 4 resolutions in order to improve the capture range and robustness of the registration.
The method uses a 4D (3D+time) free-form
B-spline deformation model and a similarity metric that minimizes variance of intensities under the constraint that the average deformation over images is zero. This constraint defines a true mean frame of reference that lie in the center of the population without having to calculate it explicitly.\footnote{\url{https://readthedocs.org/projects/simpleelastix/downloads/pdf/latest/}}

\item
\textbf{Advanced Normalizing Tools (ANTs)} is a widely-used software package which is state-of-the-art for medical image registration~\citep{avants2009advanced}.
We use the ``Rigid'' and ``Affine'' implementation for the rigid and affine model, respectively.
The volumes are registered successively at three different resolutions: 0.25x, 0.5x and finally at full resolution. At 0.25x and 0.5x resolution, Gaussian smoothing with~$\sigma$ of two and one voxels is applied, respectively. 
For non-linear registration, we use ``SyN'', which performs Symmetric Normalization~\citep{avants2008syn}. 
Finally, we used mutual information as the similarity metric for all models, which is suitable for registering images with different contrasts.

\item
\textbf{SynthMorph} is a deep learning-based registration method which achieves agnosticism to modality/contrast by leveraging a generative strategy for synthesizing diverse images, thereby supporting multi-modal registration~\citep{hoffmann2022synthmorph}.
SynthMorph accepts as input the moving and fixed images and outputs a dense deformation field instead of global affine parameters, which is a common strategy in many well-performing registration models~\citep{balakrishnan2019voxelmorph}.

Note that an important pre-processing step required by SynthMorph is an affine-registration step to a pre-defined reference space~\citep{Reuter2010HighlyAI,fischl2012freesurfer}.
Thus, this limits the flexibility of SynthMorph in the sense that all registrations are performed in this reference space. 
In contrast, {\modelnames} enables the user to define any arbitrary reference space via the fixed image.
In addition, this requirement increases pre-processing time.
In our experiments, we first affine register every image using  ANTs (see above).
SynthMorph models are implemented in Keras/Tensorflow.

\item
\textbf{EasyReg} is a deep learning-based registration tool that supports both affine and nonlinear registration in a single pipeline~\citep{iglesias2023easyreg}. 
EasyReg is designed to be symmetric, diffeomorphic, agnostic to MRI modality and resolution (in a manner similar to SynthMorph), and does not require any preprocessing or parameter tuning.

The nonlinear registration part of EasyReg is perhaps most similar to SynthMorph, where the fixed and moving image are accepted as input to a network and the output is a dense deformation field.
Unlike SynthMorph, EasyReg does not require an affine-registration step and instead performs it as an additional step in its registration pipeline. 
In particular, EasyReg uses a pretrained segmentation model to generate a parcellation of the brain, and keypoints are derived by computing the centroids of all 97 ROIs of the parcellation.
These 97 keypoints are extracted for both the moving and fixed image, and the affine transformation is solved that aligns these keypoints corresponding to Eqn.~\eqref{eq:closeform}.
Note that this affine step crucially depends on the performance and robustness of the segmentation model.

\end{itemize}

\section{Results}
\subsection{Main Results}
\subsubsection{Pairwise Registration}

We analyze the performance of baselines and our proposed \modelnames~under conditions of large initial misalignments in terms of rotation.
Figs.~\ref{fig:lineplot_harddice_vs_rot_ss_nss_unimodal_multimodal} and~\ref{fig:lineplot_hausd_vs_rot_ss_nss_unimodal_multimodal} plot overall Dice and HD across rotation angle of the moving image for baselines and \modelnames.
Each panel depicts rigid, affine, and nonlinear registrations, respectively.
Each separate figure varies different combinations of unimodal vs. multimodal, skullstripped and non-skullstripped.

Generally, {\modelnames} performs well across all rotation angles, modalities, and (non-)skullstripped.
\modelnames-L outperforms all baselines for rigid and affine transformations across all rotation angles.
However, for the case of nonlinear unimodal and multimodal skullstripped registration at small initial misalignment (e.g. 0 degrees of rotation, (bottom left panels) ANTs, SynthMorph, and EasyReg outperform \modelnames{}.  
We attribute this to the fact that TPS in 3D is likely not sufficient to align cortical geometry and therefore the model does not seem to identify cortical keypoints.
This leads to relatively worse alignment in cortical regions, which baselines (which do not have the TPS assumption) can align better.

We find that all baseline models suffer substantially as the rotation angle increases, across all transformation types. 
For classical methods (ITK-Elastix and ANTs), the models fail to find correspondences when they are far apart on the spatial grid.
For SynthMorph, the performance drop is likely due to not affine aligning the image pair, which is a requirement for SynthMorph but an added step in the pipeline.
EasyReg accounts for this by performing an initial affine alignment, but the performance drop is due to the failure of the pretrained segmentation model (used to extract ROI centroids to be used as keypoints) to segment brains which are significantly rotated, leading to a poor initial affine alignment (see Appendix).

We compare the computational time across different models in Table~\ref{tab:timing}. 
Some representative examples for all models are provided in Fig.~\ref{fig:imshow_pairwise} for qualitative evaluation.
Overall, \modelnames{} outperforms other baselines at high degrees of initial misalignment, and furthermore performs comparably or often better (at large misalignments) than the state-of-the-art ANTs registration, while requiring substantially less runtime.

\subsubsection{Groupwise Registration}

Fig.~\ref{fig:lineplot_dice_time_vs_groupsize} depicts groupwise performance and timings for {\modelnames} vs. ITK-Elastix.
We use B-spline for ITK-Elastix and TPS with $\lambda=0$ for {\modelnames}.
We find that {\modelnames} has much better and more stable performance across all group sizes we tested.
Note that on our CPU, ITK-Elastix failed on 128 subjects.
{\modelnames} has an advantage in that keypoints can be precomputed in a serial fashion on a GPU, thus enabling much better scaling to large group sizes.

Fig.~\ref{fig:imshow_groupwise} depicts a representative example of groupwise registration with 4 subjects.
The first row depicts the initial unaligned images, and the second and third row show the results of ITK-Elastix and {\modelnames}, respectively.
The last column depicts the average template brain for all 4 subjects.
We find that {\modelnames} groupwise registration to be substantially better, as evidenced by the sharp lines in the template brain.
On a GPU, {\modelnames} is also faster than ITK-Elastix by nearly 4 orders of magnitude.

\subsubsection{Longitudinal Registration}

Fig~\ref{fig:boxplot_longitudinal} shows boxplots for longitudinal registration performance across rotation angles for skull-stripped and non-skull-stripped, respectively.
We find that across all rotation angles and with and without skullstripping, {\modelnames} outperforms ITK-Elastix.

\subsubsection{Registration with Lesions}

Fig~\ref{fig:boxplot_lesion_affine} shows a boxplot of performance of all models on lesion data.
We observe that the weighted variant of {\modelnames} outperforms all baselines across most rotation angles and is generally has more stable performance. 
In particular, weighted {\modelnames} tends to outperform the unweighted variant of {\modelnames} at 45 degrees of rotation and above.

Note that we use Dice performance as a proxy for registration quality, even though the SynthSeg-generated segmentations are not guaranteed to be robust to diseased patients.

\subsection{Keypoint Analysis}
\label{sec:experiment3}

\subsubsection{Visualizing keypoints} 
In contrast to existing models that compute the transformation parameters using a “black-box” neural network, one can investigate the keypoints that \modelnames~learns to drive the alignment. 
Fig.~\ref{fig:imshow_points_ss_and_nss} shows the keypoints for a moving and fixed subject pair via mid-slices for sagittal, axial, and coronal views.  
The first three columns depict keypoints extracted from skull-stripped images, and the last three columns depict keypoints extracted from non-skull-stripped images.
The color of the keypoints represents depth with respect to the mid-slice.
The ``Aligned" slices show both warped (dots) and fixed (crosses) points.

Note that keypoint locations are trained end-to-end without explicit annotations.
We observe that keypoint locations are generally in sub-cortical regions, where anatomical variability is relatively low across subjects as compared to cortical regions.

\subsubsection{Number of Keypoints}\label{sec:num_keypoints}

\begin{figure*}[t!]
    \centering
    \begin{subfigure}[t]{0.50\textwidth}
        \centering
        \includegraphics[height=1.75in]{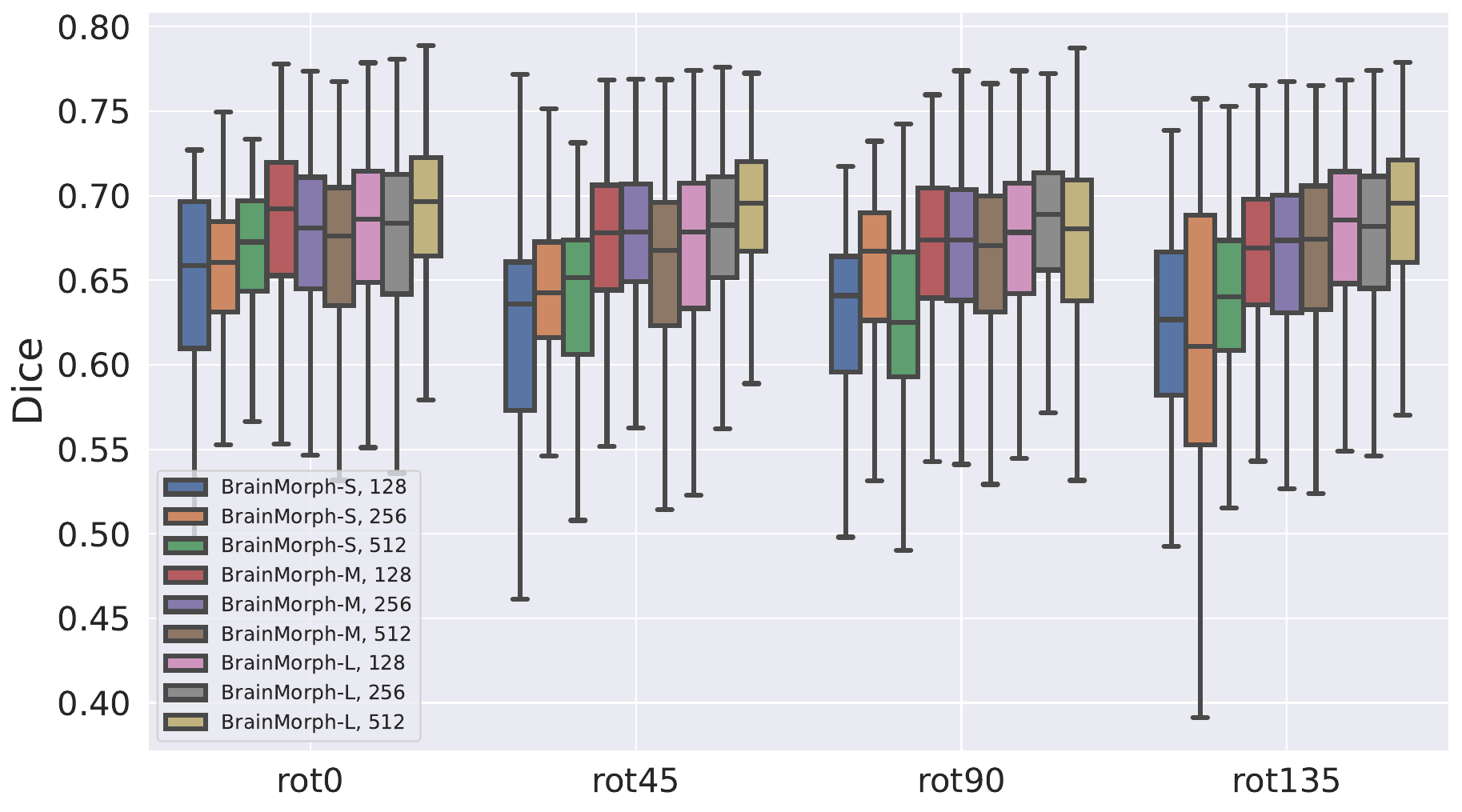}
        \caption{}
        \label{fig:keymorph_keypoint_comparison_pairwise_unimodal_skull-stripped}
    \end{subfigure}%
    ~ 
    \begin{subfigure}[t]{0.50\textwidth}
        \centering
        \includegraphics[height=1.75in]{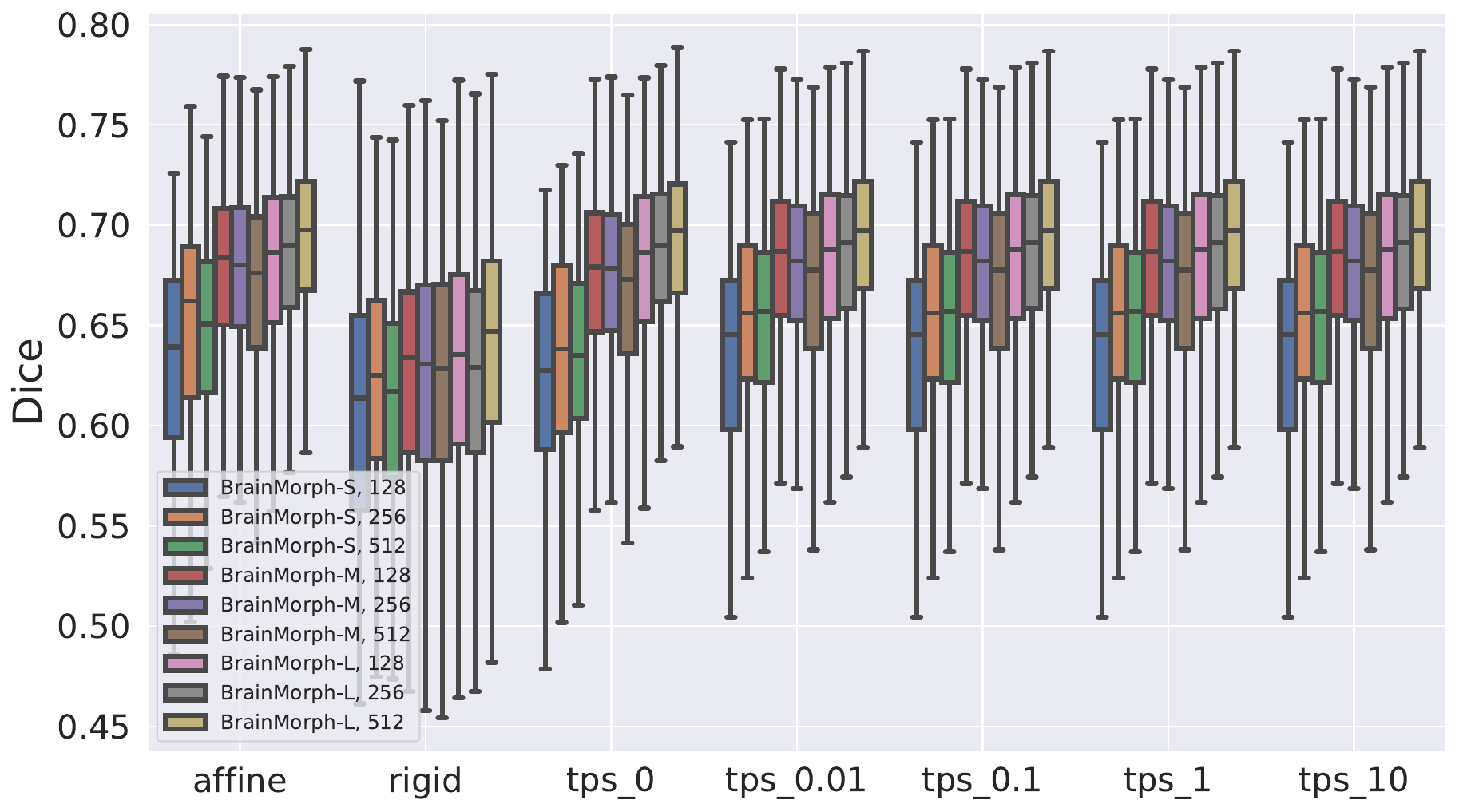}
        \caption{}
        \label{fig:boxplot_keypoints_transforms}
    \end{subfigure}
    \caption{Registration performance of variants of \modelnames{}, varying the number of keypoints a) across rotation angles and b) across transform types.}
\end{figure*}
As an ablation, we examine the effect of the number of keypoints used for alignment across different transformations.
We trained \modelnames~model variants with 128, 256, and 512 keypoints. 
Fig.~\ref{fig:keymorph_keypoint_comparison_pairwise_unimodal_skull-stripped} illustrates that performance is not discernably correlated to increasing the number of keypoints.
We hypothesize that as keypoints are generally in subcortical regions which are anatomically stable, increasing the number of correspondences does not provide further advantage beyond a certain point.
Fig.~\ref{fig:boxplot_keypoints_transforms} breaks it down across different transform types, from which we can observe that performance is relatively stable across affine and different TPS transformations.
We further hypothesize that due to the locations of keypoints, increasing the degree of nonlinearity doesn't significantly lead to improved performance.

\section{Discussion}
The results demonstrate that {\modelnames} is a robust and flexible tool for brain MRI registration.
On pairwise registration, {\modelnames} is generally superior to baselines across all degrees of initial misalignment, and is state-of-the-art for affine and rigid transformations.
These results hold consistently for for unimodal and multimodal registration, as well as skull-stripped and non-skull-stripped data.
In addition, {\modelnames} does not require extensive pre-processing like skullstripping and pre-affine registration. 
On longitudinal and groupwise registration, {\modelnames} is superior to baseline registration algorithms, while being much more memory efficient and nearly 4 orders of magnitude faster.

The main limitation of {\modelnames} is nonlinear performance at low initial misalignment and skull-stripped data, for which ANTs and SynthMorph perform excellently.
For this reason, users who require good nonlinear registrations may consider using {\modelnames} as a robust initial alignment tool, and further performing nonlinear registration using a tool like ANTs.

\section{Conclusion}
We presented a robust and flexible registration tool based on the \modelname{} framework, called \modelnames{}, which is a deep learning-based image registration method that uses corresponding keypoints to derive the optimal transformation that align the images.
This formulation enables interpretability, robustness to large initial misalignments, and flexibility/controllability of registrations at test-time.
Training on a massive dataset of over 100,000 unique images from nearly 16,000 subjects enables our tool to work on raw data with minimal pre-processing.
Empirically, we demonstrate fast and competitive performance across rigid, affine, nonlinear, and groupwise registration, particularly at large degrees of initial misalignment.

\acks{%
We thank Benjamin Billot and Juan Eugenio Iglesias for helpful discussions. Funding for this project was in part provided by the NIH grants R01AG053949, R01AG064027 and R01AG070988, and the NSF CAREER 1748377 grant.
}

%
\ethics{The work follows appropriate ethical standards in conducting research and writing the manuscript, following all applicable laws and regulations regarding treatment of animals or human subjects.}

\coi{We declare we do not have conflicts of interest.}

\data{All data supporting the findings of the study are publicly available. All references to data are included in the Appendix.
}

\bibliography{refs}


\clearpage
\appendix
\section{Details on the Keypoint Detection Network}
The keypoint detector in BrainMorph is a UNet-style architecture, but that is truncated at the last upsampling layer.
Thus, keypoints are extracted at half resolution; this is done for balancing quality of extracted keypoints with computational feasibility.

The UNet is composed of $d$ number of levels (i.e. downsampling layers), with 2 convolutional blocks per level.
$d = \{4, 5, 6\}$ corresponds to BrainMorph-\{S, M, L\}, respectively.
Each level has its corresponding block in the upsampling part of the UNet, except for the final resolution.
Each convolutional block consists of a convolutional layer, an instance normalization, and a ReLU nonlinearity.
The number of channels starts at 32 and doubles with every level.

\section{Differentiable, Closed-Form Coordinate Transformations}\label{sec:transformations}
\bpara{Notation:} In the following sections, column vectors are lower-case bolded and matrices are upper-case bolded. 
$D$-dimensional coordinates are represented as column vectors, i.e. $\bm{p} \in \mathbb{R}^D$. $D$ is typically 2 or 3. 
$\tilde{\bm{p}}$ denotes $\bm{p}$ in homogeneous coordinates, i.e. $\tilde{\bm{p}} = [\bm{p}, 1]^T$.
Superscripts in parentheses $\bm{p}^{(i)}$ index over separate instances of $\bm{p}$ (e.g. in a dataset), whereas subscripts $\bm{p}_i$ denotes the $i$'th element of $\bm{p}$.

We summarize three parametric transformation families that can be derived in closed-form, from corresponding keypoint pairs.
Suppose we have a set of $N$ corresponding keypoint pairs $\{(\bm{p}^{(i)}, \bm{q}^{(i)})\}_{i=1}^N$, where $\bm{p}^{(i)}, \bm{q}^{(i)} \in \mathbb{R}^D$ and $N > D$.
For convenience, let $\bm{P} := \begin{bmatrix}\bm{p}^{(1)} & ...  & \bm{p}^{(N)} \end{bmatrix}\in \mathbb{R}^{D\times N}$, and similarly for $\tilde{\bm{P}}$ and $\bm{Q}$. 
Define $\mathcal{T}_\theta : \mathbb{R}^D \rightarrow \mathbb{R}^D$ as a family of coordinate transformations, where $\theta \in \Theta$ are parameters of the transformation.
For all transformation families, we also consider weighted versions, where we have weights for each correspondence $\{w_i\}_{i=1}^N$.
For convenience, let $\bm{W} = \text{diag}(w_1, ..., w_N)$.

\subsection{Rigid}
Rigid transformations apply a rotation $\bm{R} \in \mathbb{R}^{D \times D}$ and a translation $\bm{t} \in \mathbb{R}^{D \times 1}$ to a coordinate:
\begin{equation}
    \mathcal{T}_\theta(\bm{p}) = \bm{R}\bm{p} + \bm{t},
\end{equation}
where the parameter set is the elements of the matrix and vector,~\mbox{$\theta = \{\bm{R}, \bm{t}\}$}.

The optimal translation is estimated by subtracting the weighted centroids of the moving and fixed point clouds:
\begin{equation}
    \bm{t}^* = \bar{\bm{p}} - \bar{\bm{q}},
\end{equation}
where $\bar{\bm{p}} = \sum_i \bm{p}^{(i)}$ for non-weighted and $\bar{\bm{p}} = \sum_i w_i \bm{p}^{(i)}$ for weighted, and similarly for $\bar{\bm{q}}$.

The optimal rotation is well-studied and is known as the orthogonal Procrustes problem~\citep{viklands2006algorithms}.
First, denote by $\bar{\bm{P}}$ the centered version of $\bm{P}$ where each column is subtracted by the centroid $\bar{\bm{p}}$, and similarly for $\bar{\bm{Q}}$. 
Next, compute the SVD of the weighted cross-correlation matrix $\text{SVD}(\bm{\Sigma}) = \text{SVD}(\bar{\bm{P}}^T \bar{\bm{Q}}) = \bm{U} \bm{\Lambda} \bm{V}^T$.
For weighted, $\Sigma = \bar{\bm{P}}^T \bm{W} \bar{\bm{Q}}$.
Then, $\bm{R}^* = \bm{V}\bm{U}^T$.


\subsection{Affine}
Affine transformations are represented as a matrix multiplication of $\bm{A} \in \mathbb{R}^{D \times (D+1)}$ with a coordinate in homogeneous form:
\begin{equation}
    \mathcal{T}_\theta(\bm{p}) = \bm{A}\tilde{\bm{p}},
\end{equation}
where the parameter set is the elements of the matrix,~\mbox{$\theta = \{\bm{A}\}$}.

Given $N$ corresponding keypoint pairs, there exists a differentiable, closed-form expression for an affine transformation that aligns the keypoints:
\begin{align} 
\opttran(\bm{P}, \bm{Q}) &:= \argmin_{\theta} \sum_{i=1}^N \left(\bm{A} \tilde{\bm{p}}^{(i)} - \bm{q}^{(i)}\right)^2 \\
    &= \bm{Q}\tilde{\bm{P}}^{T}(\tilde{\bm{P}}\tilde{\bm{P}}^{T})^{-1}.
\label{eq:closeform}
\end{align}

To derive this solution, rewrite the objective in matrix form:
\begin{align*} 
\mathcal{L} &= \sum_{i=1}^N \left(\bm{A} \tilde{\bm{p}}^{(i)} - \bm{q}^{(i)}\right)^2 \\
            &=\left \| \bm{A} \tilde{\boldsymbol{P}}- \boldsymbol{Q}\right \|_{F},
\end{align*}
where~$\norm{\cdot}_F$ denotes the Frobenius norm.
Taking the derivative with respect to~$\bm{A}$ and setting the result to zero, we obtain:
\begin{align*}
    \frac{\partial \mathcal{L}}{\partial \bm{A}} &=  (\bm{A}\tilde{\boldsymbol{P}} - \boldsymbol{Q}) \tilde{\boldsymbol{P}}^{T} = \boldsymbol{0} \\
    &\implies  \bm{A}\tilde{\boldsymbol{P}}\tilde{\boldsymbol{P}}^{T}   = \boldsymbol{Q}\tilde{\boldsymbol{P}}^{T} \\
    &\implies  \bm{A} = \boldsymbol{Q}\tilde{\boldsymbol{P}}^{T} (\tilde{\boldsymbol{P}}\tilde{\boldsymbol{P}}^{T})^{-1}.
\end{align*}

The extension to incorporate weightings for the correspondences is straightforward:
\begin{align} 
\opttran(\bm{P}, \bm{Q}, \bm{W}) &:= \argmin_{\theta} \sum_{i=1}^N w_i \left(\bm{A} \tilde{\bm{p}}^{(i)} - \bm{q}^{(i)}\right)^2 \\
    &= \bm{Q}\bm{W}\tilde{\bm{P}}^{T}(\tilde{\bm{P}}\bm{W}\tilde{\bm{P}}^{T})^{-1}.
\end{align}
%

Note that solving for the affine transformation is the least-squares solution to an overdetermined system, and thus in practice the points will not be exactly matched due to the restrictive nature of the affine transformation.
This restrictiveness may be alleviated or removed by choosing a transformation family with additional degrees of freedom, as we detail next.

\subsection{Thin-Plate Spline}
The application of the thin-plate spline (TPS) interpolant to modeling coordinate transformations yields a parametric, non-rigid deformation model which admits a closed-form expression for the solution that interpolates a set of corresponding keypoints~\citep{bookstein1989tps,donato2002approximatetps,rohr2001landmark,machado2018ultrasoundsurgery,frisken2020brainshift}.
This provides additional degrees of freedom over the affine family of transformations, while also subsuming it as a special case. 

For the $d$'th dimension, the TPS interpolant $\mathcal{T}_{\theta_d} : \mathbb{R}^D \rightarrow \mathbb{R}$ takes the following form:
\begin{equation}
    \mathcal{T}_{\theta_d}(\bm{p}) = (\bm{a}_d)^T \tilde{\bm{p}} + \sum_{i=1}^N v_{i,d} U\left(\norm{\bm{p}^{(i)} - \bm{p}}_2\right),
\end{equation}
where $\bm{a}_d \in \mathbb{R}^{D+1}$ and $\{v_{i,d}\}$ constitute the transformation parameters $\theta_d$ and $U(r) = r^2 \ln(r)$. 
We define $\bm{A} \in \mathbb{R}^{(D+1) \times D}$ and $\bm{V} \in \mathbb{R}^{N \times D}$ as the collection of all the parameters for $d=1, ..., D$.
Then, the parameter set is $\theta = \{\bm{A}, \bm{V}\}$.


This form of $\mathcal{T}$ minimizes the \textit{bending energy}:
\begin{equation}
    I_{\mathcal{T}} = \int_{\mathbb{R}^D} \norm{\nabla^2 \mathcal{T}}_F^2 d\bm{p}_1...d\bm{p}_D,
\end{equation}
where~$\norm{\cdot}_F$ is the Frobenius norm and $\nabla^2 \mathcal{T}$ is the matrix of second-order partial derivatives of $\mathcal{T}$. 
For each $\theta_d$, we impose interpolation conditions $\mathcal{T}_{\theta_d}(\bm{p}^{(i)}) = \bm{q}_d^{(i)}$ and enforce $\mathcal{T}$ to have square-integrable second derivatives: 
\begin{equation}
    \sum_{i=1}^N v_{i,d} = 0 \ \ \text{and} \ \ 
    \sum_{i=1}^N v_{i,d} \bm{p}_d = 0 \ \ \forall d \in \{1, ..., D\}.
\end{equation}
Given these conditions, the following system of linear equations solves for $\theta$: 
\begin{equation}
    \bm{\Psi} \theta :=
    \begin{bmatrix}
    \bm{K} & \bm{L} \\
    \bm{L}^T & \bm{O}
    \end{bmatrix}
    \begin{bmatrix}
    \bm{V} \\
    \bm{A}
    \end{bmatrix}
    =
    \begin{bmatrix}
    \bm{Q}^T \\
    \bm{O}
    \end{bmatrix}
    := \bm{Z}.
    \label{eq:block-tps}
\end{equation}
Here, $\bm{K} \in \mathbb{R}^{N\times N}$ where $\bm{K}_{ij} = U\left(\norm{\bm{p}^{(i)} - \bm{p}^{(j)}}_2\right)$, $\bm{L} \in \mathbb{R}^{N \times (D+1)}$ where the $i$'th row is $(\tilde{\bm{p}}^{(i)})^T$, and $\bm{O}$ is a matrix of zeros with appropriate dimensions.
Thus,
\begin{equation}
    \opttran(\bm{P}, \bm{Q}) := \bm{\Psi}^{-1} \bm{Z}.
\end{equation}
Solving for $\theta^*$ is a differentiable operation.

The interpolation conditions can be relaxed (e.g. under the presence of noise) by introducing a regularization term:
\begin{equation}
    \argmin_{\theta_d} \sum_{i=1}^N \left(\mathcal{T}_{\theta_d}\left(\bm{p}^{(i)}\right) - \bm{q}_d^{(i)}\right)^2 + \lambda I_{\mathcal{T}}
\end{equation}
where $\lambda>0$ is a hyperparameter that controls the strength of regularization.
As $\lambda$ approaches $\infty$, the optimal $\mathcal{T}$ approaches the affine case (i.e. zero bending energy). 
This formulation can be solved exactly by replacing $\bm{K}$ with $\bm{K}+\lambda \bm{I}$ in Eq.~\eqref{eq:block-tps}.
Importantly, $\theta$ and the optimal $\theta^*(\bm{P}, \bm{Q})$ exhibits a dependence on $\lambda$.
Finally, weights can be incorporated by replacing $\bm{K}$ with $\bm{K}+\lambda \bm{W}^{-1}$~\citep{rohr2001landmark}.

\section{List of Datasets}
A full list of datasets used for training and evaluation is shown in Table~\ref{tab:gigamed_citations_in_captions}.
\begin{sidewaystable*}[!ht]
    \centering
    \fontsize{7}{8}\selectfont  
    \begin{tabular}{lllllllcc}
    \toprule
        \textit{Dataset} & \textit{Description} & \textit{Train/Eval} & \textit{No. Subjects} & \textit{No. Modalities} & \textit{Modalities Info} & \textit{No. Images} & \textit{Longitudinal?} & \textit{Lesion?} \\ \midrule
        \rowcolorone
        BraTS-SSA-2023 & African Gliloms & Train & 60 & 4 & T1 T1gd T2 FLAIR & 240 & \xmark & \cmark \\ 
        BraTS-MEN-2023 & Meningiomas & Train & 1000 & 4 & T1 T1gd T2 FLAIR & 4000 & \xmark & \cmark \\ 
        \rowcolorone
        BraTS-MET-2023 & Metastases & Train & 165 & 4 & T1 T1gd T2 FLAIR & 660 & \xmark & \cmark \\ 
        BraTS-MET-NYU-2023 & Metastases & Train & 164 & 4 & T1 T1gd T2 FLAIR & 656 & \xmark & \cmark \\ 
        \rowcolorone
        BraTS-PED-2023 & Pediatric Glioblastomas & Train & 99 & 4 & T1 T1gd T2 FLAIR & 396 & \xmark & \cmark \\ 
        BraTS-MET-UCSF-2023 & Metastases & Train & 324 & 4 & T1 T1gd T2 FLAIR & 1296 & \xmark & \cmark \\ 
        \rowcolorone
        BraTS-2016 & Tumor & Train & 1104 & 1 & T1 & 1104 & \xmark & \cmark \\ 
        UCSF-BMSR & Metastases & Train & 459 & 4 & FLAIR T1post T1pre subtraction & 1836 & \xmark & \cmark \\ 
        \rowcolorone
        MSSEG2 & Multiple Sclerosis & Train & 40 & 1 & FLAIR & 80 & \cmark & \cmark \\ 
        Shifts-challenge-part1MSSEG & Multiple Sclerosis & Train & 15 & 5 & FLAIR T1 T2 GADO DP & 75 & \xmark & \cmark \\ 
        \rowcolorone
        Shifts-challenge-part2Ljubljana & Multiple Sclerosis & Train & 25 & 4 & FLAIR T1 T2 T1ce & 100 & \xmark & \cmark \\ 
        Shifts-challenge-part2Best & Multiple Sclerosis & Train & 21 & 4 & FLAIR T1 T2 PD & 84 & \xmark & \cmark \\ 
        \rowcolorone
        Brain Development & Adult and Neonatal Brain Atlases  & Train & 33 & 1 & T1 & 33 & \xmark & \xmark \\ 
        EPISURG & Epileptic & Train & 133 & 1 & T1 & 133 & \xmark & \xmark \\ 
        \rowcolorone
        WMH & White Matter Hyperintensity & Train & 120 & 1 & T1 & 120 & \xmark & \xmark \\ 
        ISLES-2022 & Ischemic Stroke Lesion & Train & 250 & 2 & DWI ADC & 500 & \xmark & \cmark \\ 
        \rowcolorone
        MedicalDecathlon-BrainTumour & Tumor & Train & 484 & 4 & FLAIR T1 T1gd T2 & 1936 & \xmark & \cmark \\ 
        UCSF-ALPTDG & Post Treatment Gliomas & Train & 298 & 5 & FLAIR T1 T1ce T2 T1csub & 1490 & \cmark & \cmark \\ 
        \rowcolorone
        StanfordMETShare & Metastases & Train & 105 & 4 & T1 FLAIR T1gd T1pre & 420 & \xmark & \cmark \\ 
        PPMI & Parkinson's Disease & Train & 1382 & 1 & T1 & 1382 & \cmark & \xmark \\ 
        \rowcolorone
        ADNI & Alzheimer's Disease & Train & 2027 & 3 & T1 T2 FLAIR & 18645 & \cmark & \xmark \\ 
        PAC 2019 & Normal Aging & Train & 2640 & 1 & T1 & 2640 & \xmark & \xmark \\ 
        \rowcolorone
        AIBL & Alzheimer's Disease & Train & 708 & 1 & T1 & 708 & \xmark & \xmark \\ 
        OASIS1 & Normal Aging and Dementia & Train & 348 & 1 & T1 & 1350 & \cmark & \xmark \\ 
        \rowcolorone
        OASIS3 & Normal Aging and Alzheimer’s Disease & Train & 2270 & 6 & T1 T2 T2star FLASH FLAIR angio & 10299 & \cmark & \xmark \\ 
        IXI & Normal & Eval & 151 & 3 & T1 T2 PD & 453 & \xmark & \xmark \\ 
        \rowcolorone
        OASIS2 & Dementia & Eval & 298 & 1 & T1 & 1095 & \cmark & \xmark \\ 
        BraTS-GLI-2023 (test) & Glioblastomas & Eval & 876 & 4 & T1 T1gd T2 FLAIR & 3504 & \xmark & \cmark \\ 
        \rowcolorone
        MS-60 & Multiple sclerosis & Eval & 60 & 3 & T1 T2 FLAIR & 180 & \xmark & \cmark \\ 
        \midrule 
        \textbf{Total} & ~ & ~ & \textbf{15659} & \textbf{13} (unique) & ~ & \textbf{55415} ($\times$ 2) & ~ & ~ \\ \bottomrule
    \end{tabular}
    \caption{Dataset details. Total images is 55415 (possibly non-unique due to overlap). $\times$ 2 = 110836 to include with and without skullstripping. All citations in top-bottom row order:~\citep{kazerooni2024brats}, \citep{kazerooni2024brats},~\citep{kazerooni2024brats},~\citep{kazerooni2024brats},~\citep{kazerooni2024brats},~\citep{rudie2024metsucsf},~\citep{menze2015brats},~\citep{rudie2024metsucsf},~\citep{commowick2018MSLesionSegmentation},~\citep{malinin2022shifts},~\citep{rudie2024metsucsf},~\citep{rudie2024metsucsf},~\citep{Gousias2012NewbornMRI},~\citep{perez2020episurg},~\citep{Kuijf2019WMHSegmentation},~\citep{HernandezPetzsche2022ISLES},~\citep{antonelli2022decathlon},~\citep{rudie2022ucsfalptg},~\citep{Grovik2020BrainMetastasesMRI},~\citep{marek2011ppmi},~\citep{Petersen2010ADNI},~\citep{Fowler2021AIBLStudy},~\citep{fisch2021pac},~\citep{marcus2007oasis1},~\citep{lamontagne2019oasis3},~\url{https://brain-development.org/ixi-dataset/},~\citep{marcus2010oasis2},~\citep{kazerooni2024brats},~\citep{Muslim2022msdataset}.}\label{tab:gigamed_citations_in_captions}
\end{sidewaystable*}

\section{Failure Cases of Baselines}
\begin{figure*}[t]
\centering
\includegraphics[width=0.9\linewidth]{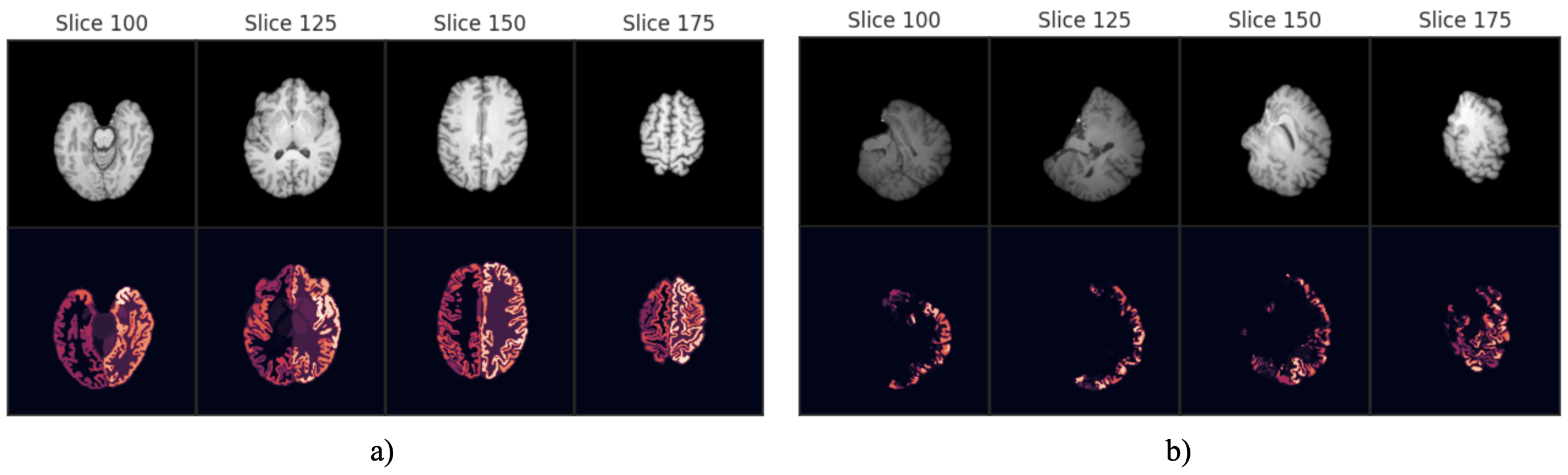}
   \caption{EasyReg fails to segment images that are significantly rotated. a) non-rotated brain, b) rotated brain.}
\label{fig:easyreg_segmentations}
\end{figure*}

We find that all baseline models suffer substantially as the rotation angle increases, across all transformation types. 
For classical methods (ITK-Elastix and ANTs), the models fail to find correspondences when they are far apart on the spatial grid.
For SynthMorph, the performance drop is likely due to not affine aligning the image pair, which is a requirement for SynthMorph but an added step in the pipeline.
EasyReg accounts for this by performing an initial affine alignment, but the performance drop is due to the failure of the pretrained segmentation model (used to extract ROI centroids to be used as keypoints) to segment brains which are significantly rotated, leading to a poor initial affine alignment.
We visualize this in Fig.~\ref{fig:easyreg_segmentations}.

\begin{figure*}[h!]
\centering
\includegraphics[width=\linewidth]{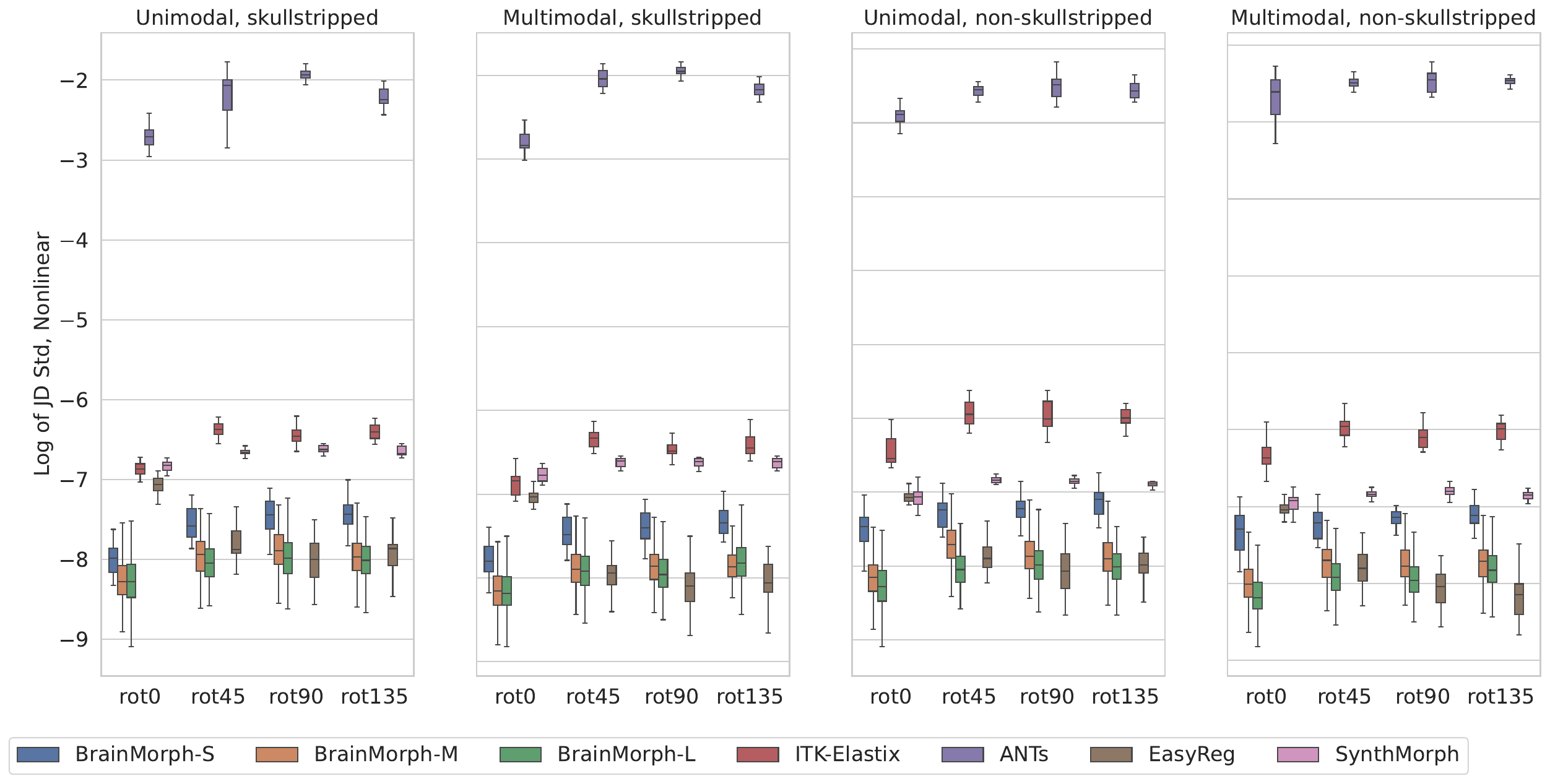}
\caption{Log of standard deviation of Jacobian determinant on pairwise registration. Unimodal/multimodal, skull-stripped/non-skull-stripped.}
\label{fig:boxplot_jdstd_vs_rot_ss_nss_unimodal_multimodal}
\end{figure*}

\end{document}